\documentclass[a4paper]{article}
\usepackage{geometry}
\usepackage{tikz}
\usepackage{cite}
\usetikzlibrary{calc}
\usetikzlibrary{arrows,shapes,chains}
\usepackage{xcolor}
\usepackage{amsfonts}
\usepackage{amsmath}
\usepackage{graphicx}
\usepackage{multirow}
\usepackage{algorithmic}
\usepackage{algorithm}
\usepackage{subfigure}
\usepackage{epstopdf}
\usepackage{url}
\DeclareMathOperator*{\argmax}{arg\,max}
\DeclareMathOperator*{\argmin}{arg\,min}

\newcommand{\cb}[1]{\ifmmode {\boldsymbol{#1}}\else ${\boldsymbol{#1}}$\fi}
\newcommand{\cp}[1]{\ifmmode {\mathcal{#1}}\else ${\mathcal{#1}}$\fi}
\newcommand{\bx}{\cb{x}}
\newcommand{\bz}{\cb{z}}
\newcommand{\bc}{\cb{c}}

\newcommand{\by}{\cb{y}}

\title{Spectral-Spatial Feature Extraction and Classification by ANN Supervised with Center Loss \\in Hyperspectral Imagery}
\author{Alan J.X. Guo and Fei~Zhu
	\thanks{A.~Guo and F.~Zhu are with the Center for Applied Mathematics, Tianjin University, China. (jiaxiang.guo;~fei.zhu@tju.edu.cn)}
}

\begin{document}
	
\maketitle
\begin{abstract}
In this paper, we propose a spectral-spatial feature extraction and classification framework based on artificial neuron network (ANN) in the context of hyperspectral imagery.
With limited labeled samples, only spectral information is exploited for training and spatial context is integrated posteriorly at the testing stage.
Taking advantage of recent advances in face recognition, a joint supervision symbol that combines softmax loss and center loss is adopted to train the proposed network, by which intra-class features are gathered while inter-class variations are enlarged. Based on the learned architecture, the extracted spectrum-based features are classified by a center classifier.
Moreover, to fuse the spectral and spatial information, an adaptive spectral-spatial center classifier is developed, where multiscale neighborhoods are considered simultaneously, and the final label is determined using an adaptive voting strategy.
Finally, experimental results on three well-known datasets validate the effectiveness of the proposed methods compared with the state-of-the-art approaches. 		
\end{abstract}
	
\section{Introduction}
Recent advances in remote sensing technology lead to improved imaging quality in terms of both spectral and spatial resolutions for hyperspectral imagery~\cite{fauvel2013advances}.
Typically, a hyperspectral image is collected by remote sensors with hundreds of channels over a wavelength range. It is regarded as a data cube consisting of continuous images, each acquired by a certain channel over the same land-cover.
Consequently, every pixel corresponds to a reflectance spectrum. In hyperspectral imagery analysis, an essential task is the differentiation of target materials by identifying the class of each pixel.

To address the hyperspectral classification problem, traditional feature extraction (FE) and classification strategies have been extensively investigated, seeking to extract shallow representations for input spectra. Extracting features with linear transformations permits representing the original data in a subspace with reduced-dimension. This category of methods includes principal component analysis (PCA)~\cite{licciardi2012linear}, independent component analysis (ICA)~\cite{villa2011hyperspectral} and linear discriminant analysis (LDA)~\cite{bandos2009classification}, to name a few. Since the nonlinearities often exist in hyperspectral images, various nonlinear methods have been proposed accordingly, such as support vector machine (SVM)~\cite{melgani2004classification}, manifold learning~\cite{lunga2014manifold}, and the kernel-based algorithms~\cite{kuo2009kernel,scholkopf2002learning}.

Most of the aforementioned methods are spectrum-based ones, {\em i.e.,} to extract the features directly from the pixel-wise spectra while ignoring the intrinsic geographical structure in specific data. In fact, the spatially adjacent pixels tend to be inherently correlated. A joint usage of the spectral and spatial information could improve the representation ability of the extracted features \cite{fauvel2013advances}. To this end, numbers of spectral-spacial methods have been derived and have outperformed the spectral ones in terms of classification accuracy~\cite{chen2011hyperspectral,fauvel2008spectral,huo2011spectral,li2013spectral,plaza2009incorporation,song2014remotely}. In~\cite{Fang2014Spectral}, the multiscale adaptive sparse representation (MASR) method exploits the spatial information derived from neighborhoods of varying sizes by an adaptive sparse strategy. Despite promising classification results achieved on several datasets, MASR produces high computational cost, since multiscale windows around a testing pixel are constructed, in which every pixel needs to be sparsely coded using a structural dictionary consisting of training samples. To effectively use the spectral-spatial information remains a challenging problem.
Meanwhile, the greatly improved resolutions in both spectral and spatial domains may enlarge the intra-class variation ({\em e.g.}, roofs with shadows) and decrease the interclass variation ({\em e.g.}, roads and roofs have similar spectra), thus greatly affecting the classification accuracy~\cite{zhao2016spectral}.

Deep learning has emerged as a powerful feature extraction technology and gained great success in many machine learning tasks. Compared with traditional methods, deep models are capable to extract more abstract and complex features with better representation ability.
The features are generated in a hierarchical manner, namely the high-level representations are successfully built from a cascade of lower layers consisting of nonlinear processing units.
However, to train a deep network with large number of unknown parameters, a large-scale training set is required. Otherwise, the network is prone to overfitting. This becomes a major obstacle to exploiting deep learning in hyperspectral imagery, where the number of labeled samples are quite limited due to high expense of manually labeling and even the available labels are not always reliable~\cite{Jiao2017Deep}.
To extract deep spectral-spatial representation, a conventional strategy is to train a network or a classifier based on patch-based samples, as did in~\cite{chen2016deep, liang2016hyperspectral, slavkovikj2015hyperspectral,zhao2016spectral,Jiao2017Deep}. However, this will further aggregate the sample shortage problem, if the overlap of training and testing samples is not allowed.

To alleviate this dilemma with deep models, small networks with respectively less parameters have been advocated, including the earlier stacked autoencoder (SAE)~\cite{benediktsson2005classification,chen2014deep}, the deep belief network (DBN)~\cite{chen2015spectral}, and a few models based on
the convolutional neural networks (CNN). In~\cite{romero2016unsupervised}, an unsupervised CNN has been proposed to obtain sparse representations, mainly by using the greedy layerwise unsupervised pre-training along with the efficient enforcing of population and lifetime sparsity.
Three supervised CNN architectures, referred by $1$-D, $2$-D and $3$-D CNNs, were investigated in~\cite{chen2016deep}. More specifically, $3$-D CNN could jointly extract the spectral-spatial integrated features by working on data cube composed by a neighborhood of spectra. The authors enlarged the training set by virtual sample enhancement, and avoided the overfitting issue by a few techniques, such as dropout and $\ell_2$-norm regularization. Similar strategies were also adopted in a novel CNN architecture proposed in~\cite{yu2017convolutional}. In~\cite{zhao2016spectral}, a spectral-spatial feature
based classification framework was proposed. To address the increasing intra-class variation and interclass confusion issues in spectral domain, the spectral feature in low-dimensionality was firstly extracted by the so-called BLDE method, aiming to increase the interclass variation while keeping the intra-class samples stay close. The deep spatial-based feature was extracted by CNN. Finally, the fusion feature, stacked by extracted spectral and spatial features, was used to train the classifier.

Taking advantage of pre-training and fine-tuning techniques, relative deep networks have also been investigated in the literature. The pre-training on large-scale data sets other than target data is advisable, since the initial layers of deep network tend to be generic filters, such as edge or color blob detector, which are independent from specific data and transferable for different recognition tasks~\cite{Jiao2017Deep}. Studies in~\cite{Nogueira2017Towards} compared three possible practices to exploit existing CNNs, namely full-training, fine-tuning, and using CNNs as feature extractor, finding out that fine-tuning tended to be the best strategy.   
In~\cite{Jiao2017Deep}, a deep multiscale spatial-spectral feature extraction algorithm is proposed to learn effective discriminant features of hyperspectral images. The VGG-verydeep-16 net is firstly pre-trained on the natural image data sets, and then used to extract the spatial structural feature. The fusion vector stacked by the spatial and spectral features is fed to the classifier.

To explore spectral-spatial information, the routine by training with patch-based pixels might easily cause the overlap between training and testing sets. An alternative is postprocessing of the extracted features at the testing stage. Authors in~\cite{li2017hyperspectral} proposed a deep pixel-pair feature (PPF) classification framework based on a well-designed CNN. Pixel-pairs are generated by taking Cartesian product and used to train a deep model. For a testing sample, paired samples are generated by combining the centering pixel with its neighbors and classified using the learned network. The final label is predicted by a voting strategy. The idea of postprocessing will also be adopted in this paper, namely spatial context is only exploited at the testing stage. Therefore, PPF and the aforementioned MASR are chosen for comparison in Section~\ref{sec: Experiments}, since these two state-of-the-art methods are trained only based on spectral samples.

Raised in the field of face recognition, deep metric learning is becoming a hotspot in deep learning and has gained great success. The basic idea is to enhance the discriminative power of the deeply learned features by simultaneously enlarging the inter-class variations and decreasing the intra-class variations. To this end, several metric learning loss functions have been introduced to CNN as supervision signal, such as triplet loss~\cite{schroff2015facenet}, magnet loss~\cite{rippel2016metric}, and newly introduced center loss~\cite{wen2016discriminative} that will be investigated in this paper. To the best of our knowledge, to design a neuron network based on deep metric learning has not been approached so far in the context of hypersepctral  FE and classification.

In this paper, we propose a deep spectral-spatial FE and classification framework that allows to extract discriminative spectral-spatial features with intra-class compactness and interclass dispersity. We apply a joint supervision symbol proposed in~\cite{wen2016discriminative} to train the network, in which the conventional softmax loss and the so-called center loss is combined. The spectral pixels are directly fed to the network to estimate the network parameters and the class centers as well. Based on the learned network, the extracted features are classified by a center classifier.
To further improve the classification performance, we propose to adaptively integrate the spatial information from multi-scale neighborhoods at the testing stage. For each scale, the label and its associated weight are estimated using center classifier, and the final label is determined by a voting strategy.  
The proposed framework has following characteristics:
\begin{enumerate}
  \item \emph{Discriminative features}: Taking advantage of center loss, the proposed ANNC allows to extract spectral features that are not only separable but also discriminative. This helps to decrease the undesirable side effects brought by the high-resolutions to the classification task, namely the increasing intra-class variation and inter-class similarity. The extracted features fit the proposed center classifier well. Compared with raw spectra, the discriminative features are easier to handle, facilitating an effective integration of spatial information at the testing stage.

  \item \emph{Less overfitting issues}: Facing the limited number of training samples in hyperspectral images, we design a small network with less parameters in order to alleviate the overfitting issue. To make full use of training set, virtual samples are generated. Moreover, only spectral information is exploited at the training stage to avoid overlap of training and testing sets. As indicated in~\cite{wen2016discriminative}, the joint supervision which complicates the training task also helps to alleviate the overfitting problem to some degree.
  \item \emph{Training with less prior knowledge on the image}: In the proposed framework, the training procedure is strictly spectrum-based, without requiring any local information on the training pixel, such as the coordinate and the (unlabeled) neighboring pixels that may confront the testing set. This offers a more flexible model for real world applications with less conditions to satisfy.
\end{enumerate}

The remainder of this paper is structured as follows. Section~\ref{Sec:Background} succinctly describes the basis of artificial neural network. In Section~\ref{sec:ProposedModel}, we propose a deep spectral-spatial FE and classification framework. Experimental results and analysis on three real hyperspectral images are presented in~\ref{sec: Experiments}. Finally, Section~\ref{sec: Conclusion} provides some conclusion remarks.

\section{Basic concepts of ANN}\label{Sec:Background}
\begin{figure}[ht]
	\centering
        \scriptsize
        \tikzstyle{format}=[circle,draw,thin,fill=white]
        \begin{tikzpicture}[node distance=5mm,  auto,>=latex',  thin,  start chain=going below, every join/.style={norm}, scale=0.45]
            \definecolor{hidden_color}{RGB}{101,123,131}
            \definecolor{input_color}{RGB}{181,137,0}
            \definecolor{output_color}{RGB}{42,161,152}
            \filldraw[fill=input_color,rounded corners,fill opacity=0.3,style=dashed] (2,1) rectangle (8,3);
            \filldraw[fill=hidden_color,rounded corners,fill opacity=0.3,style=dashed] (0,4) rectangle (10,9);
            \filldraw[fill=output_color,rounded corners,fill opacity=0.3,style=dashed] (2,10) rectangle (8,12);

            \node at (3,2) [format] (v11){$v_{11}$};
            \node at (7,2) [format] (v12){$v_{12}$};
            \node at (1,5) [format] (v21){$v_{21}$};
            \node at (5,5) [format] (v22){$v_{22}$};
            \node at (9,5) [format] (v23){$v_{23}$};
            \node at (1,8) [format] (v31){$v_{31}$};
            \node at (5,8) [format] (v32){$v_{32}$};
            \node at (9,8) [format] (v33){$v_{33}$};
            \node at (5,11) [format](v41){$v_{41}$};

            \draw[->] (v11) -- (v21);
            \draw[->] (v11) -- (v21);
            \draw[->] (v11) -- (v23);
            \draw[->] (v12) -- (v21);
            \draw[->] (v12) -- (v22);
            \draw[->] (v12) -- (v23);

            \draw[->] (v21) -- (v31);
            \draw[->] (v21) -- (v32);
            \draw[->] (v21) -- (v33);
            \draw[->] (v22) -- (v31);
            \draw[->] (v22) -- (v32);
            \draw[->] (v22) -- (v33);
            \draw[->] (v23) -- (v31);
            \draw[->] (v23) -- (v32);
            \draw[->] (v23) -- (v33);

            \draw[->] (v31) -- (v41);
            \draw[->] (v32) -- (v41);
            \draw[->] (v33) -- (v41);

            \node at (-3,2) [scale=0.6]{\large Input layer};
            \node at (-3,6.5) [scale=0.6]{\large Hidden layers};
            \node at (-3,11) [scale=0.6]{\large Output layer};
        \end{tikzpicture}
        \caption{\label{instance_nn} An example of ANN with $2$ hidden layers.}
\end{figure}
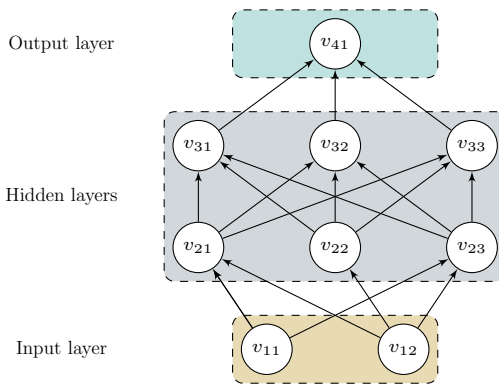

Artificial neural network (ANN) is inspired by the organization and functioning of biological neurons~\cite{wang2003artificial,Hill1994Artificial}. Typically, it is viewed as a trainable multilayer structure with multiple feature extraction steps, and is composed by an input layer of neurons, several hidden layers of neurons, and a final layer of output neurons, as shown in~\figurename~\ref{instance_nn}.

Given an ANN architecture, let $v_{ij}$ denote the $j$-th neuron in the $i$-th layer. This value is computed based on the neurons in previous layer, and takes the form
\begin{equation}\label{eq:ANN}
	v_{ij} = \Phi \left( b_{ij}+\sum_{k = 1}^{n_{i-1}} w_{(i-1)k}^{ij}v_{(i-1)k}\right),
\end{equation}
where $n_{i-1}$ denotes the number of neurons in the $(i-1)$-th layer, $w_{(i-1)k}^{ij}$ represents the weight connecting neurons $v_{ij}$ and $v_{(i-1)k}$, and $b_{ij}$ indexes the bias for the neuron $v_{ij}$. 
The pointwise activation function $\Phi(\cdot)$ is often applied to introduce nonlinearity to the neutron network. In this paper, the commonly-used ReLU function is chosen, which is defined by $$\Phi(x) = \max(0, x).$$

The loss function measures the global error between the ANN output and the groundtruth. To train a network, namely to determine the parameters of weight and bias, one seeks to minimize the loss function defined on the given training set consisting of labeled samples in pair. The resulting optimization problem is addressed by the celebrated gradient-based, error back-propagation algorithm. 

To avoid gradient computation on the whole training set, the stochastic gradient descent (SGD) with mini-batch strategy is often applied, which alleviates the computational burden by processing a mini-batch at each iteration of training.

The main objective of dropout is to regularize the neuron network in order to prevent overfitting~\cite{Dropout}. During the training step, the weights associated to some units are excluded from being updated temporarily, and these units are randomly chosen with a fixed probability, namely dropout ratio.

\section{Proposed spectral-spatial feature extraction and classification framework}\label{sec:ProposedModel}
This section firstly presents the proposed neuron network, followed by explaining the joint supervision symbol used for training. Based on the learned architecture, a spectral center classifier is proposed at the testing stage. Finally, a spectral-spatial FE and classification framework termed ANNC-ASSCC is presented.
\subsection{Proposed ANN structure}\label{subsec:Structure}
\begin{figure}
        \centering
		\scriptsize
		\tikzstyle{format}=[circle,draw,thin,fill=white]
		\tikzstyle{format_gray}=[circle,draw,thin,fill=gray]
		\tikzstyle{format_rect}=[rectangle,draw,thin,fill=white,align=center]
	    \tikzstyle{arrowstyle} = [->,thick]
        \scalebox{1}
        {
		\begin{tikzpicture}[node distance=4mm,  auto,>=latex',  thin,  start chain=going below, every join/.style={norm}, scale=0.3]
		\definecolor{gray_so}{RGB}{88,110,117}
		\definecolor{yellow_so}{RGB}{181,137,0}
		\definecolor{cyan_so}{RGB}{42,161,152}
		\definecolor{orange_so}{RGB}{203,75,22}
		\filldraw[fill=yellow_so,rounded corners,fill opacity=0.33,style=dashed] (-2,-2) rectangle (2,15);
		\foreach \x/\xtext in {0, 2, 4, 6, 8, 10, 12, 14, 16, 18}
		{
			\node at (0,\x/4+3.75) [format] { };
		    \node at (0,\x/4+0.25+3.75) [format_gray] { };
	    }
        \filldraw[fill=orange_so,rounded corners,fill opacity=0.2,style=dashed] (3,-2) rectangle (17,15);
        \foreach \x/\xtext in {0,2,4,6,8,10,12,14,16,18,20,22,24,26,28,30,32,34,36,38,40,42,44,46,48}
        {
        	\node at (5,\x/4) [format] { };
        	\node at (5,\x/4+0.25) [format_gray] { };
        }
        \filldraw[fill=cyan_so,rounded corners,fill opacity=0.33,style=dashed] (18,-2) rectangle (27,15);
		\foreach \x/\xtext in {0,2,4,6,8,10,12,14,16,18,20,22,24,26,28}
		{
			\node at (10,\x/4+2.5) [format] { };
			\node at (10,\x/4+0.25+2.5) [format_gray] { };
		}
		\foreach \x/\xtext in {0, 2, 4, 6, 8, 10, 12}
	    {
	    	\node at (15,\x/4+4.5) [format] { };
	    	\node at (15,\x/4+0.25+4.5) [format_gray] { };
	    }
        \foreach \x/\xtext in {0, 2, 4, 6}
        {
    	    \node at (20,\x/4+5) [format_gray] { };
    	    \node at (20,\x/4+0.25+5) [format] { };
        }
        \node at (20,8/4+5) [format_gray] { };

        \foreach \x/\xtext in {0, 2, 4, 6}
        {
        	\node at (25,\x/4+5) [format_gray] { };
        	\node at (25,\x/4+0.25+5) [format] { };
        }
        \node at (25,8/4+5) [format_gray] { };

		\draw[arrowstyle] (1,6.25) -- node {FC} (4,6.25);
		\draw[arrowstyle] (6,6.25) --  node {FC} (9,6.25);
		\draw[arrowstyle] (11,6.25) -- node {FC} (14,6.25);
		\draw[arrowstyle] (16,6.25) -- node {FC} (19,6.25);
		\draw[arrowstyle] (21,6.25) -- node {Softmax} (24,6.25);
	    \node at (0,0) (n0) {};
	    \node at (5,0) (n5) {};
	    \node at (10,0) (n10) {};
	    \node at (15,0) (n15) {};
	    \node at (20,0) (n20) {};
	    \node at (25,0) (n25) {};
        \node[below of =n0] (i) {Input};
        \node[below of =n5] (l1) {Layer 1};
        \node[below of =n10] (l2) {Layer 2};
        \node[below of =n15] (l3) {Layer 3};
        \node[below of =n20] (o) {Output};
        \node[below of =n25] (label) {Label};
        \node at (15,12) [format_rect] (centerloss) {center loss};
        \node at (25,12) [format_rect] (loss) {softmax loss};
        \draw[arrowstyle] (15,8.5) to (centerloss.south);
        \draw[arrowstyle] (25,7.75) to (loss.south);
        \node at (10,14) [rounded corners,style=dashed,color=gray_so] (input_section) {Feature Extraction Network};
        \node at (22.5,14) [rounded corners,style=dashed,color=gray_so] (input_section) {Logistic Regression};
		\end{tikzpicture}
}
		\caption{\label{struct} Structure of the proposed neural network.}
\end{figure}
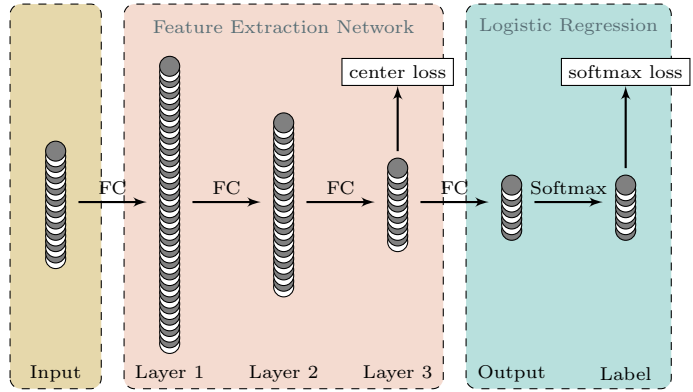

We first briefly describe the proposed ANN architecture. As illustrated in~\figurename~\ref{struct}, the proposed network consists of an input layer, three hidden layers, and an output layer followed by a softmax classifier, and all the layers are fully connected. We call the $i$-th layer fully connected, if every neuron in it has connections to all activations in the previous layer, namely all the parameters $w_{(i-1)k}^{ij}$ in~\eqref{eq:ANN} are set learnable.

The sizes of the input and output layers are determined according to the hyperspectral image under study: the former is set as the size of each sample pixel $L$, and the later is chosen to be the number of classes $K$. The numbers of neurons in Layer $1,2$ and $3$ are empirically set to be $512,256$ and $32$.

Layer $1$ and $2$ consist of  ReLU as the activation function, while Layer $3$ does not transformed with any activation function.
The output layer is activated by the softmax function in order to perform the classification task based on the learned features, namely the outputs of Layer 3.

It is noteworthy that the pixel-wise spectra are directly fed into the proposed network at the training stage, differing from most existing spatial-spectral deep learning models which are trained using patch-based samples, each being a neighborhood area centering at every labeled pixel~\cite{chen2016deep, liang2016hyperspectral, slavkovikj2015hyperspectral}.

\subsection{Joint supervision of softmax loss and center loss}\label{subsec:CenterL}
Inspired by a discriminative feature learning strategy proposed for deep face recognition in~\cite{wen2016discriminative}, we propose to train the aforementioned network by simultaneously supervising the conventional softmax loss and the newly-introduced center loss. 

Suppose that the proposed network is optimized by the SGD with mini-batch strategy at the training stage, namely the network parameters are updated based on a small set of samples at each iteration. Let $\{(\bx_1,y_1),\ldots,(\bx_M,y_M)\}$ be a given mini-batch composed by $M$ labeled samples, where $\bx_i \in \mathbb{R}^{L \times 1}$ denotes the $i$-th input spectra in the batch, and $y_i \in \{1,...,K\}$ is the corresponding ground-truth label. 

{
The softmax loss combines softmax function with multinomial logistic loss. Softmax function is commonly used in multi-class classification, as its output could be regarded as a probability distribution over predefined classes. Let $\sigma(\bz)=[\sigma_1(\bz),\sigma_2(\bz),\ldots,\sigma_K(\bz)]^{\top}$ represent the softmax function over the $K$-dimensional output feature
$\bz=[z_1,z_2,\ldots,z_K]^{\top}$. The predicted probability that sample $\bx_i$ belongs to class $j$ is given by
\begin{equation}~\label{eq:Prob}
	p(\hat{y}_i=j|\bx_i) = \sigma_j(\bz) = \frac{\exp({z_j})}{\sum_{l=1}^{K}{\exp({z_l})}},
\end{equation}
where $\hat{y}_i$ is the predicted label of sample $\bx_i$.
Multinomial logistic loss generalizes the logistic loss from binary classification to multi-class case, and is formulated as
\begin{equation}~\label{eq:Softmax}
	\mathcal{L}_S = -\frac{1}{M}\left[\sum_{i=1}^{M}\sum_{j=1}^{K}\mathbb{I}_{\{j\}}(y_i)\log{p(\hat{y}_i=j|\bx_i)}\right],
\end{equation}
where
 $\mathbb{I}_{\mathcal{A}}(x)$ is
the indicator function of the set $\mathcal{A}$ defined by
\begin{equation*}
	\mathbb{I}_{\mathcal{A}}(x) =
	\left\{
    	\begin{array}{ll}
	         1 &\forall x \in \mathcal{A}; \\
             0 &\textrm{otherwise.}
    	\end{array}
	\right.
\end{equation*}
By integrating the expression~\eqref{eq:Prob} to~\eqref{eq:Softmax}, we get the expression of the softmax loss.
}

In~\cite{wen2016discriminative}, the center loss was firstly introduced to address the face recognition task, and has achieved state-of-the-art performance on several benchmarks. Intuitively, it is designed to enhance the separability of the deeply learned features, by minimizing the intra-class distances while encouraging the inter-class variations. To this end, the center loss is defined by averaging the distances between every extracted feature and the corresponding class center, namely
\begin{equation}\label{center_loss}
	\mathcal{L}_C = \frac{1}{2M}\sum_{i=1}^{M}||\hat{\bx}_i-\bc_{y_i}||_2^2.
\end{equation}
Here, the learned feature $\hat{\bx}_i$ corresponds to the $i$-th input spectra in the batch, for $i=1,...,M$, and $\bc_k$ denotes the $k$-th class center defined by the averaging over the features in the $k$-th class, for $k=1,...,K$.
In practice, the class centers are updated with respect to mini-batch instead of being updated based on the entire training set. At the $t$-th training round, let $\bc_k^{t}$ denote the $k$-th estimated class center, and let $\bar{\bc}_k^{t}$ denote the averaged features of the $i$-th class based on mini-batch. Following~\cite{wen2016discriminative}, the class centers are iteratively updated at every training round, and are formulated as
\begin{equation}
    \bc_k^{t+1} = \bc_k^{t} + \alpha (\bar{{\bc}}_{k}^{ t+1}-\bc_k^{t}),
\end{equation}
with $\bc_k^{1} = {\bar{\bc}}_k^{1}$, for $k=1,...,K$.
Here, a hyperparameter $\alpha$ is introduced to control the learning rate of the centers, thus avoiding large perturbations. In this paper, we set $\alpha=0.5$.

The loss function of the proposed architecture is then defined by combining the softmax loss $\mathcal{L}_S$ and the center loss $\mathcal{L}_C$, with
\begin{equation}
    \mathcal{L} = \mathcal{L}_S + \lambda \mathcal{L}_C,
\end{equation}
where hyperparameter $\lambda$ balances the importance between two loss functions, and is fixed to be $0.01$ in this paper.

We apply the algorithm given in~\cite{wen2016discriminative} to train the proposed network. See Algorithm~\ref{alg:train} and find more details in the related paper.  
\begin{algorithm}[h]
	\caption{Training algorithm~\cite{wen2016discriminative}}\label{alg:train}
	\begin{algorithmic}[1]
		\renewcommand{\algorithmicrequire}{\textbf{Input:}}
		\renewcommand{\algorithmicensure}{\textbf{Output:}}
		\REQUIRE The number of iteration $t\gets 0$. Mini-batches $\{(\bx_1^t,y_1^t),\ldots,(\bx_M^t,y_M^t)\}$ for training.
		Initialized parameters $\theta$ in the network . 
		Hyperparameters $\lambda, \alpha$ and learning rate $\mu^t.$
		\ENSURE The trained network parameters $\theta$.
		\STATE Initialize the parameters $\theta$ by Gaussian distribution $N(0,0.01^2)$ (weight parameters) and constant number
		$0$ (bias parameters).
		\STATE Compute the joint loss by $\mathcal{L}^0 = \mathcal{L}_S^0 + \lambda \mathcal{L}_C^0$.
		\STATE Compute the backpropagation error $\frac{\partial \mathcal{L}^0}{\partial \bx_i^0}$ for each $i$ by
		$\frac{\partial \mathcal{L}^0}{\partial \bx_i^0} = \frac{\partial \mathcal{L}_S^0}{\partial \bx_i^0}
		+\lambda \frac{\partial \mathcal{L}_C^0}{\partial \bx_i^0}.$
		\STATE  Update the parameters $\theta$ by $\theta^0=\theta -
		\mu^0 \sum_i \frac{\partial \mathcal{L}^0}{\partial \bx_i^0}\frac{\partial \bx_i^0}{\partial \theta}$.
		\STATE Update the estimated centers by $\bc_k^0 = \bc_k'^0$, for $1\le k \le K$.
		\WHILE {not converge}
	        \STATE $t \gets t+1.$
	    	\STATE Compute the joint loss $\mathcal{L}^t = \mathcal{L}_S^t + \lambda \mathcal{L}_C^t$
	    	\STATE Compute the backpropagation error $\frac{\partial \mathcal{L}^t}{\partial \bx_i^t}$ for each $i$, by
    		$\frac{\partial \mathcal{L}^t}{\partial \bx_i^t} = \frac{\partial \mathcal{L}_S^t}{\partial \bx_i^t}
    		+\lambda \frac{\partial \mathcal{L}_C^t}{\partial \bx_i^t}.$
    		\STATE Update the parameters $\theta$ by $\theta^t=\theta^{t-1} -
    		\mu^t \sum_i \frac{\partial \mathcal{L}^t}{\partial \bx_i^t}\frac{\partial \bx_i^t}{\partial \theta^{t-1}}.$
	    	\STATE Update the estimated centers $\bc_k^t = \bc_k^{t-1}+\alpha (\bc_k'^t-\bc_k^{t-1})$, for $1\le k \le K$.
     \ENDWHILE
	\end{algorithmic}
\end{algorithm}

\subsection{Discriminative spectral feature extraction and center classifier}
\label{subsec:ANNC-SCC}
At the testing stage, the spectral features are firstly generated as the outputs of Layer 3, using the learned network. Taking advantage of center loss, the obtained features have good discriminative ability for subsequent classification task, namely the features are distributed compactly within class while different classes stay separable.

To perform classification based on the extracted features, we introduce the so-called center classifier, which assigns label to a given feature according to its nearest class center. As the real class centers are not accessible, they are approximated by averaging the learned features for training samples from each class. Let $\widehat{\bc}_k$ denote the $k$-th approximate class center, it is expressed as
\begin{equation}
	\hat{\bc}_k = \frac{\sum_{\bx \in \{\bx_{i}|y_{i}=k \}} f(\bx)}{\mathrm{card} \{\bx_{i}|y_{i}=k\}},
\end{equation}
for $1\le k \le K$, where $\{(\bx_{i},y_{i})\}$ represents the entire training set composed by pairs of spectrum and associated groundtruth label, and $f(\cdot)$ returns to the learned feature (the outputs of Layer 3) for any entry.
For a testing sample ${\bx}$, it is classified to have label $k_0$ such that
\begin{equation}
	k_0 = \argmin_{k}\{d(f(\bx),\hat{\bc}_k)\},
\end{equation}
where $d(\cdot,\cdot)$ measures the Euclidean distance between two vectors.

The above testing stage is combined with the training stage presented in Section~\ref{subsec:CenterL}, and the whole framework is termed artificial neuron network with center loss and spectral center classifier (ANNC-SCC). For comparison, the case without center loss is abbreviated by ANN-SCC.
\subsection{Spectral-spatial feature generation and classification}
\label{subsec:ANNC-ASSCC} 
Existing works in~\cite{chen2016deep, liang2016hyperspectral, slavkovikj2015hyperspectral,zhao2016spectral,Jiao2017Deep} seek the spectral-spatial representation by training a model with patch-based samples. Although effective, this may further aggregate the sample shortage problem in hyperspectral classification. Thanks to center loss, the spectral features extracted using the proposed framework have good enough discriminative ability that enables an effective integration of spatial information posteriorly at the testing stage. 

1) \emph{Spectral-Spatial Center Classifier}: We propose a spectral-spatial center classifier by applying center classifier to the averaged spectral feature, which is calculated within a neighborhood around a testing sample.
Firstly, the spectral features, {\em i.e.}, the outputs of Layer $3$, of the whole image are extracted and stored. Given a testing sample $\bx$, let $N_{i\times i}(\bx)$ denote a local neighborhood of samples centering at $\bx$, where ${i\times i}$ represents the pre-defined neighborhood scale with $i$ being an odd number. The average feature of the neighborhood, represented by $\tilde{f}(\bx)$ is computed and taken as the spectral-spatial feature for the testing sample $\bx$, as summarized in Algorithm~\ref{ave_feature}. Here, all the training samples are excluded and not taken into account. The new feature is fed to the aforementioned center classifier.
The method is termed artificial neuron network with center-loss and spectral-spatial center classifier (ANNC-SSCC).

\begin{algorithm}
	\caption{Spatial-spectral feature generation }\label{ave_feature}
	\renewcommand{\algorithmicrequire}{\textbf{Input:}}
	\renewcommand{\algorithmicensure}{\textbf{Output:}}
	\begin{algorithmic}[1]
		\REQUIRE The testing sample $x$. The neighborhood of samples $N_{i\times i}(\bx)$ centering at $\bx$ (with $\bx$ included).
		The training set $T$.
		\ENSURE The spectral-spatial feature $\tilde{f}(\bx)$
		\STATE Initialize the summation variable $\mathrm{featureSum} = 0$ and counter $\mathrm{cnt} = 0$.
		\FORALL {$\bz$ in neighborhood $N_{i\times i}(x)$}
		\IF {$z$ not in training set $T$}
		\STATE $\mathrm{featureSum} = \mathrm{featureSum} + f(\bz)$;
               $\mathrm{cnt} = \mathrm{cnt} + 1$.
		\ENDIF
		\ENDFOR
		\RETURN $\tilde{f}(\bx) = \frac{\mathrm{featureSum}}{\mathrm{cnt}}$
	\end{algorithmic}
\end{algorithm}
2) \emph{Adaptive Spectral-Spatial Center Classifier}:
In ANNC-SSCC, the neighborhood scale greatly influences the classification performance and should be appropriately set according to the feature distribution and local contextual property. However, to obtain such information is not practical, thus posing an obstacle to the estimation of neighborhood scale. To overcome this difficulty, we improve ANNC-SSCC by adopting an adaptive-weighted multiscale voting strategy, where neighborhoods of varying scales are considered simultaneously.

For a testing sample $\bx$, let \{$\tilde{f}_{i\times i}(\bx)$, $\ell_{i\times i}(\bx)$\} denote the pair of spectral-spatial feature and predicted label computed by ANNC-SSCC, and let $P=\mathrm{card}(\{\ell_{i\times i}(\bx)\})$ be the number of candidate labels. Eight scales of neighborhoods, {\em i.e.}, $3 \times 3, 5 \times 5,...,17 \times 17 $ are taken into account.
Considering that smaller distance between a feature to its predicted class center means more reliable label prediction, the weight of neighborhood $N_{i\times i}$ is estimated by
\begin{equation}
w_{i\times i (\bx)} = \frac{1}{d(\tilde{f}_{i\times i}(\bx), \hat{\bc}_{\ell_{i\times i}(\bx)})},
\end{equation}
where $d(\cdot,\cdot)$ is Euclidean distance between the feature and its predicted class center,
for $i=3,5,..,17$. After gathering the neighborhoods with the same label, the weight for candidate label $p$ is expressed by
\begin{equation}
W_p(\bx)=\sum_{\{i\mid \ell_{i\times i}(\bx)=p\}}w_{i\times i},
\end{equation}
for $p=1,...,P$.
The sample $\bx$ is predicted to have label $p_0$, such that
\begin{equation}
{p_0}= \argmax_{p}\{W_p(\bx)\}.
\end{equation}

\begin{figure*}
    \centering
	\tikzstyle{format}=[circle,draw,thin,fill=white]
	\tikzstyle{format_gray}=[circle,draw,thin,fill=gray]
	\tikzstyle{format_rect}=[rectangle,draw,thin,fill=white,align=center]
	\tikzstyle{arrowstyle} = [->,thick]
	\tikzstyle{network} = [rectangle, minimum width = 3cm, minimum height = 1cm, text centered, draw = black,align=center,rounded corners,fill=green_so,fill opacity=0.5,text opacity=1]
	\tikzstyle{training_batch} = [trapezium, trapezium left angle = 30, trapezium right angle = 150, minimum width = 3cm, text centered, draw = black, fill = cyan_so, fill opacity=0.3,text opacity=1,align=center]		
	\tikzstyle{class_features} = [trapezium, trapezium left angle = 30, trapezium right angle = 150, minimum width = 3cm, text centered, draw = black, fill = cyan_so, fill opacity=0.3,text opacity=1,align=center]
	\tikzstyle{pixel} = [rectangle, draw = black, fill = orange_so, fill opacity=0.3,text opacity=0,align=center]	
	\tikzstyle{feature} = [rectangle, draw = black, fill = orange_so, fill opacity=0.3,text opacity=0,align=center,rounded corners]			
	\tikzstyle{arrow1} = [thick, ->, >= stealth]
	\tikzstyle{arrow2} = [thick, dashed, ->, >= stealth]
	\begin{tikzpicture}[auto,>=latex',  thin,  start chain=going below, every join/.style={norm}]
	\definecolor{gray_so}{RGB}{88,110,117}
	\definecolor{yellow_so}{RGB}{181,137,0}
	\definecolor{cyan_so}{RGB}{42,161,152}
	\definecolor{orange_so}{RGB}{203,75,22}
	\definecolor{green_so}{RGB}{133,153,0}
	\definecolor{red_so}{RGB}{220,50,47}
	\definecolor{magenta_so}{RGB}{211,54,130}
	\definecolor{violet_so}{RGB}{108,113,196}
	\useasboundingbox (0,0) rectangle (20*0.77,10*0.77);
	\scope[transform canvas={scale=0.77}]
		\coordinate (zero) at (0,-1);
	    \filldraw[fill=gray_so,rounded corners,fill opacity=0.1,style=dashed] ($(zero)+(0,8)$) rectangle ($(zero)+(5,10.5)$);
    	\node(b1) at ($(zero)+(2.5,10)$)[training_batch] {Data \& labels $\#1$};
    	\node(b2) at ($(b1)-(0,0.5)$)[training_batch] {Data \& labels $\#2$};
    	\node(b3) at ($(b2)-(0,0.5)$)[training_batch,text opacity=0,fill opacity=0.1,style=dashed, draw opacity=0.7] {Data \& labels $\#m$};
    	\node(dotsb3) at ($(b2)-(0,0.5)$) {$\cdots$};
    	\node(b4) at ($(b3)-(0,0.5)$)[training_batch] {Data \& labels $\#M$};
		\node(train_model) at ($($(b1)!0.5!(b4)$)+(5,0)$)[network] {Neural network \\ \scriptsize{supervised with joint loss}};
    	\draw[arrow2] ($($(b1)!0.5!(b4)$)+(2.5,0)$) -- node(a1){} (train_model);
		\node(model) at ($(train_model)+(0,-2.5)$)[network,fill=yellow_so] {Neural network \\ \scriptsize{Trained model}};
		\draw[arrow2] (train_model) -- (model);
		\node(model2) at ($(train_model)+(5,0)$)[network,fill=yellow_so] {Neural network \\ \scriptsize{Trained model}};
		\draw[arrow2] (train_model) -- (model2);
		\draw[arrow1,line width=2pt] ($(b4.south)+(0,-0.25)$) to +(0,-1.25) |- (model);
		\node(f4) at ($(b1)+(3,-7)$)[class_features,fill=violet_so]{\normalsize{Features of training data}};
		\node(class_4)[align=center] at ($(f4.west)-(1,0)$) {\\Class $K$};
		\node(f3) at ($(f4)+(0,0.5)$)[class_features,fill=violet_so,text opacity=0,fill opacity=0.1,style=dashed, draw opacity=0.7]{\normalsize{Features of training data}};
		\node(class_3)[align=center] at ($(f3.west)-(1,0)$) {\\$\ldots$};
		\node(f2) at ($(f3)+(0,0.5)$)[class_features,fill=violet_so,text opacity=0]{\normalsize{Features of training data}};
		\node(class_2)[align=center] at ($(f2.west)-(1,0)$) {\\Class $2$};
		\node(f1) at ($(f2)+(0,0.5)$)[class_features,fill=violet_so,text opacity=0]{\normalsize{Features of training data}};
		\node(class_1)[align=center] at ($(f1.west)-(1,0)$) {\\Class $1$};
		\filldraw[fill=gray_so,rounded corners,fill opacity=0.1,style=dashed] ($(class_1)-(0.75,-0.5)$) rectangle ($(f4.east)+(0.75,-0.5)$);
		\draw[arrow1,line width=2pt] (model.east) to ++(0.5,0) to ++(0,-1) to ++(-4,0) to ++(0,-0.75);
		\draw[arrow1,line width=2pt]  ($($(f1.east)!0.5!(f4.east)$)+(0.75,0)$) -- node(ave)[align=center]{average}($($(f1)!0.5!(f4)$)+(6.25,0)$);
		\draw[style=dashed] ($(zero)+(10,1)$) to ($(zero)+(10,11)$);
		\node(c1) at ($(f1)+(7.25,0)$)[feature,fill=magenta_so,text opacity=0]{N};
		\node(center_1)[align=center] at ($(c1.west)-(0.4,0)$) {$\widehat{c}_1$};
		\node(c2) at ($(c1)+(0,-0.5)$)[feature,fill=magenta_so,text opacity=0]{N};
		\node(center_2)[align=center] at ($(c2.west)-(0.4,0)$) {$\widehat{c}_2$};
		\node(c3) at ($(c2)+(0,-0.5)$)[feature,fill=magenta_so,text opacity=0,fill opacity=0.1,style=dashed, draw opacity=0.7]{N};
		\node(center_3)[align=center] at ($(c3.west)-(0.4,0)$) {$\cdots$};
		\node(c4) at ($(c3)+(0,-0.5)$)[feature,fill=magenta_so,text opacity=0]{N};
		\node(center_K)[align=center] at ($(c4.west)-(0.4,0)$) {$\widehat{c}_K$};
		\filldraw[fill=gray_so,rounded corners,fill opacity=0.1,style=dashed] ($(c1)-(1,-0.5)$) rectangle ($(c4)+(0.5,-0.5)$);
		\node at ($(c1)-(0.25,2.3)$) [rounded corners,style=dashed,color=gray_so] {Class Centers};
		\node(c_v)[format_rect,fill=gray_so,fill opacity=0.1,text opacity=1] at ($($(c1)!0.5!(c4)$)+(2.5,0)$) {Compare \\ \& Vote};
		\node(label)[format_rect,fill=red_so,fill opacity=0.3,text opacity=1] at ($($(c1)!0.5!(c4)$)+(5,0)$) {Predicted\\ label};
		\draw[arrow2,line width=3pt]  ($($(c1)!0.5!(c4)$)+(0.5,0)$) -- (c_v.west);
		\draw[arrow2,line width=3pt]  (c_v.east) -- (label.west);
		\node(p0)[pixel,fill=red_so,fill opacity=0.5] at ($(model2.east)+(1.5,0)$) {};
		\node(p1)[pixel] at ($(p0)-(0.275,0.275)$) {};
		\node(p2)[pixel] at ($(p0)-(0.275,0)$) {};
		\node(p3)[pixel] at ($(p0)-(0.275,-0.275)$) {};
		\node(p4)[pixel] at ($(p0)-(0,0.275)$) {};
		\node(p5)[pixel] at ($(p0)-(0,-0.275)$) {};
		\node(p6)[pixel,fill opacity=0,style=dashed,draw opacity=0.7] at ($(p0)-(-0.275,0.275)$) {};
		\node(p7)[pixel] at ($(p0)-(-0.275,0)$) {};
		\node(p8)[pixel] at ($(p0)-(-0.275,-0.275)$) {};
		\node at ($(p0)+(0,0.9)$) {$N_{3\times 3}$};
		
		\node(p10)[pixel,fill=red_so,fill opacity=0.5] at ($(p0)+(1.3,0)$) {};
		\node(p11)[pixel] at ($(p10)-(0.275,0.275)$) {};
		\node(p12)[pixel] at ($(p10)-(0.275,0)$) {};
		\node(p13)[pixel] at ($(p10)-(0.275,-0.275)$) {};
		\node(p14)[pixel] at ($(p10)-(0,0.275)$) {};
		\node(p15)[pixel] at ($(p10)-(0,-0.275)$) {};
		\node(p16)[pixel,fill opacity=0,style=dashed,draw opacity=0.7] at ($(p10)-(-0.275,0.275)$) {};
		\node(p17)[pixel] at ($(p10)-(-0.275,0)$) {};
		\node(p18)[pixel] at ($(p10)-(-0.275,-0.275)$) {};
		
		\node(p19)[pixel] at ($(p10)-(0.55,0.55)$) {};
		\node(p110)[pixel] at ($(p10)-(0.55,0.275)$) {};
		\node(p111)[pixel] at ($(p10)-(0.55,0)$) {};
		\node(p112)[pixel] at ($(p10)-(0.55,-0.275)$) {};
		\node(p113)[pixel] at ($(p10)-(0.55,-0.55)$) {};
		\node(p114)[pixel] at ($(p10)-(0.275,0.55)$) {};
		\node(p115)[pixel] at ($(p10)-(0.275,-0.55)$) {};
		\node(p116)[pixel] at ($(p10)-(0,0.55)$) {};
		\node(p117)[pixel] at ($(p10)-(0,-0.55)$) {};
		\node(p118)[pixel] at ($(p10)-(-0.275,0.55)$) {};
		\node(p119)[pixel] at ($(p10)-(-0.275,-0.55)$) {};
		\node(p120)[pixel] at ($(p10)-(-0.55,0.55)$) {};
		\node(p121)[pixel] at ($(p10)-(-0.55,0.275)$) {};
		\node(p122)[pixel] at ($(p10)-(-0.55,0)$) {};
		\node(p123)[pixel] at ($(p10)-(-0.55,-0.275)$) {};
		\node(p124)[pixel] at ($(p10)-(-0.55,-0.55)$) {};
		\node at ($(p10)+(0,0.9)$) {$N_{5\times 5}$};

		\node(p20) at ($(p10)+(1.1,0)$) {$\cdots$};
		\node at ($(p20)+(0.9,0)$) {$N_{17\times 17}$};
		
		\filldraw[fill=gray_so,rounded corners,fill opacity=0.1,style=dashed] ($(p0)-(0.7,1)$) rectangle ($(p0)+(4,1.2)$);
		\node at ($(p5)-(-1.7,-1.1)$) [rounded corners,style=dashed,color=gray_so] {Eight neighborhoods};
		\draw[arrow2,line width=3pt] ($(p0)-(0.7,0)$) to (model2);
		\node(fp0)[feature,fill=red_so,fill opacity=0.5] at ($(model2)+(-1.5,-2.25)$) {};
		\node(fp1)[feature] at ($(fp0)-(0.275,0.275)$) {};
		\node(fp2)[feature] at ($(fp0)-(0.275,0)$) {};
		\node(fp3)[feature] at ($(fp0)-(0.275,-0.275)$) {};
		\node(fp4)[feature] at ($(fp0)-(0,0.275)$) {};
		\node(fp5)[feature] at ($(fp0)-(0,-0.275)$) {};
		\node(fp6)[feature,fill opacity=0,style=dashed,draw opacity=0.7] at ($(fp0)-(-0.275,0.275)$) {};
		\node(fp7)[feature] at ($(fp0)-(-0.275,0)$) {};
		\node(fp8)[feature] at ($(fp0)-(-0.275,-0.275)$) {};
		\node at ($(fp0)+(0,0.9)$) {$F_{3\times 3}$};

		\node(fp10)[feature,fill=red_so,fill opacity=0.5] at ($(fp0)+(1.3,0)$) {};
		\node(fp11)[feature] at ($(fp10)-(0.275,0.275)$) {};
		\node(fp12)[feature] at ($(fp10)-(0.275,0)$) {};
		\node(fp13)[feature] at ($(fp10)-(0.275,-0.275)$) {};
		\node(fp14)[feature] at ($(fp10)-(0,0.275)$) {};
		\node(fp15)[feature] at ($(fp10)-(0,-0.275)$) {};
		\node(fp16)[feature,fill opacity=0,style=dashed,draw opacity=0.7] at ($(fp10)-(-0.275,0.275)$) {};
		\node(fp17)[feature] at ($(fp10)-(-0.275,0)$) {};
		\node(fp18)[feature] at ($(fp10)-(-0.275,-0.275)$) {};
		\node(fp19)[feature] at ($(fp10)-(0.55,0.55)$) {};
		\node(fp110)[feature] at ($(fp10)-(0.55,0.275)$) {};
		\node(fp111)[feature] at ($(fp10)-(0.55,0)$) {};
		\node(fp112)[feature] at ($(fp10)-(0.55,-0.275)$) {};
		\node(fp113)[feature] at ($(fp10)-(0.55,-0.55)$) {};
		\node(fp114)[feature] at ($(fp10)-(0.275,0.55)$) {};
		\node(fp115)[feature] at ($(fp10)-(0.275,-0.55)$) {};
		\node(fp116)[feature] at ($(fp10)-(0,0.55)$) {};
		\node(fp117)[feature] at ($(fp10)-(0,-0.55)$) {};
		\node(fp118)[feature] at ($(fp10)-(-0.275,0.55)$) {};
		\node(fp119)[feature] at ($(fp10)-(-0.275,-0.55)$) {};
		\node(fp120)[feature] at ($(fp10)-(-0.55,0.55)$) {};
		\node(fp121)[feature] at ($(fp10)-(-0.55,0.275)$) {};
		\node(fp122)[feature] at ($(fp10)-(-0.55,0)$) {};
		\node(fp123)[feature] at ($(fp10)-(-0.55,-0.275)$) {};
		\node(fp124)[feature] at ($(fp10)-(-0.55,-0.55)$) {};
		\node at ($(fp10)+(0,0.9)$) {$F_{5\times 5}$};
		
		\node(fp20) at ($(fp10)+(1.1,0)$) {$\cdots$};
		\node at ($(fp20)+(0.9,0)$) {$F_{17\times 17}$};
		
		\filldraw[fill=gray_so,rounded corners,fill opacity=0.1,style=dashed] ($(fp0)-(0.7,1)$) rectangle ($(fp0)+(4,1.2)$);
		\draw[arrow2,line width=3pt] (model2.south) to ($(model2.south)+(0,-0.5)$);
		\node at ($(fp4)-(-1.7,1)$) [rounded corners,style=dashed,color=gray_so] {Features};
		\node(fpa)[feature,fill=red,fill opacity=0.7] at ($(fp0)+(6,0)$) {N};
		\node(fpb)[feature,fill=magenta_so,fill opacity=0.7] at ($(fpa)+(0.7,0)$) {N};
		\node(fpc) at ($(fpa)+(1.4,0)$) {$\cdots$};
		\node(fpd)[feature,fill=orange_so,fill opacity=0.7] at ($(fpa)+(2.1,0)$) {N};
		\filldraw[fill=gray_so,rounded corners,fill opacity=0.1,style=dashed] ($(fpa)-(0.5,0.5)$) rectangle ($(fpd)+(0.5,0.5)$);
		\draw[arrow2,line width=3pt] ($(fp20)+(1.6,0)$) -- node(averagefeatures){average} ($(fpa)+(-0.55,0)$);
		\draw[arrow2,line width=3pt] ($(fpb)+(0.35,-0.55)$) to ($(fpb)-(-0.35,1.7)$) to ($(fpb)-(2.45,1.7)$) to (c_v.north);
		\node at ($(zero)+(5,1.5)$) [rounded corners,style=dashed,color=gray_so] {\Large{Training stage}};
		\node at ($(zero)+(15,1.5)$) [rounded corners,style=dashed,color=gray_so] {\Large{Testing stage}};
		
	\endscope
	\end{tikzpicture}	
	\caption{\label{train_test_procedure} Flowchart of the proposed ANNC-ASSCC framework. Training and testing stages are distinguished by thin and thick dashed arrows, respectively. The estimation of class centers with trained model is drawn with solid thick arrows. At the testing stage, the dashed rectangle in white represents the excluded training sample, appearing within the neighborhoods of a testing sample in red.}  style.where three arrow types are used to distinguish different procedures.
\end{figure*}
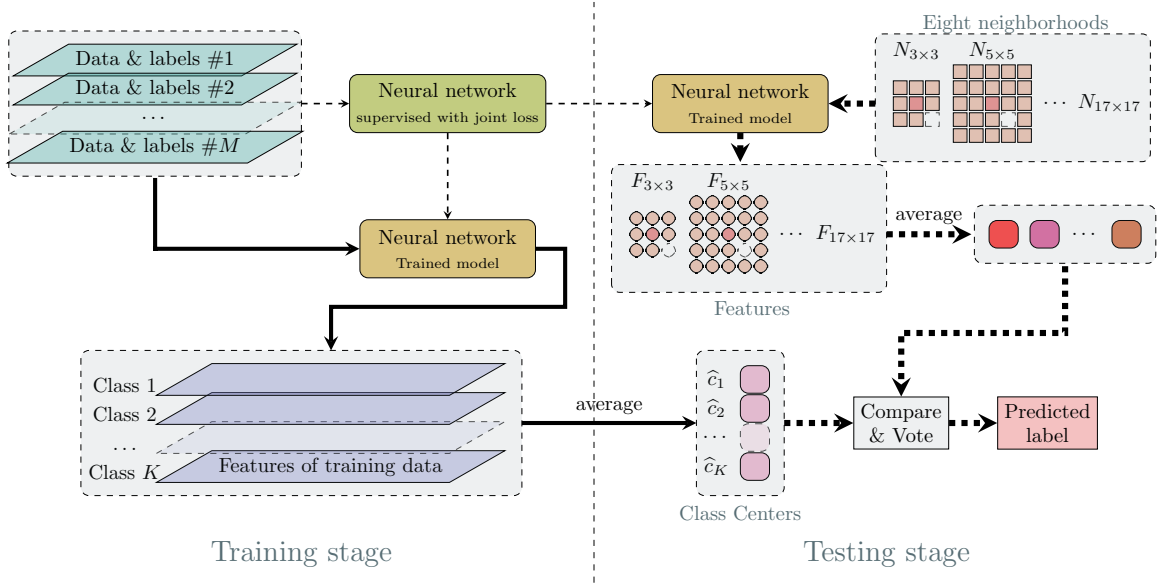

The whole training and testing procedure is referred as artificial neuron network with center-loss and adaptive spatial-spectral center classifier (ANNC-ASSCC), and the flowchart is depicted in~\figurename~\ref{train_test_procedure}.

\section{Experimental results and analysis }\label{sec: Experiments}
\subsection{Data description and quantitative metrics}
\begin{figure}[htb]
	\centering
	\includegraphics[trim =16mm 20mm 25mm 10mm, clip,width=.12\textwidth] {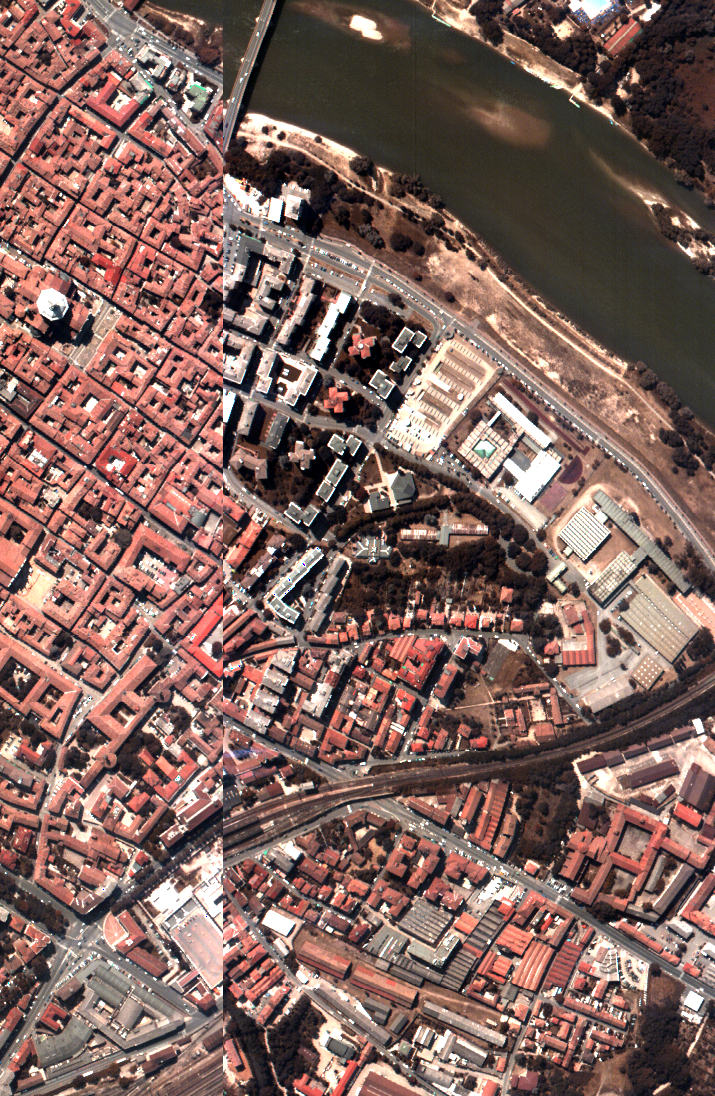}
    \includegraphics[trim =16mm 20mm 25mm 10mm, clip,width=.12\textwidth]{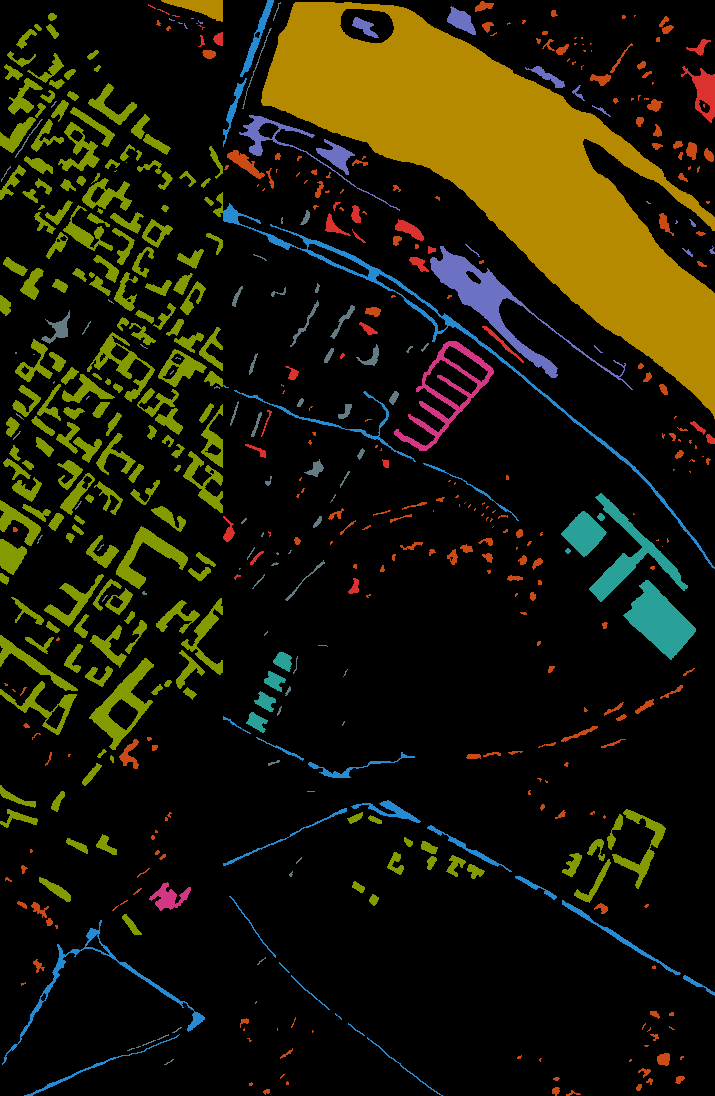}
    \includegraphics[trim =0mm 0mm 0mm 0mm, clip,width=.1\textwidth]{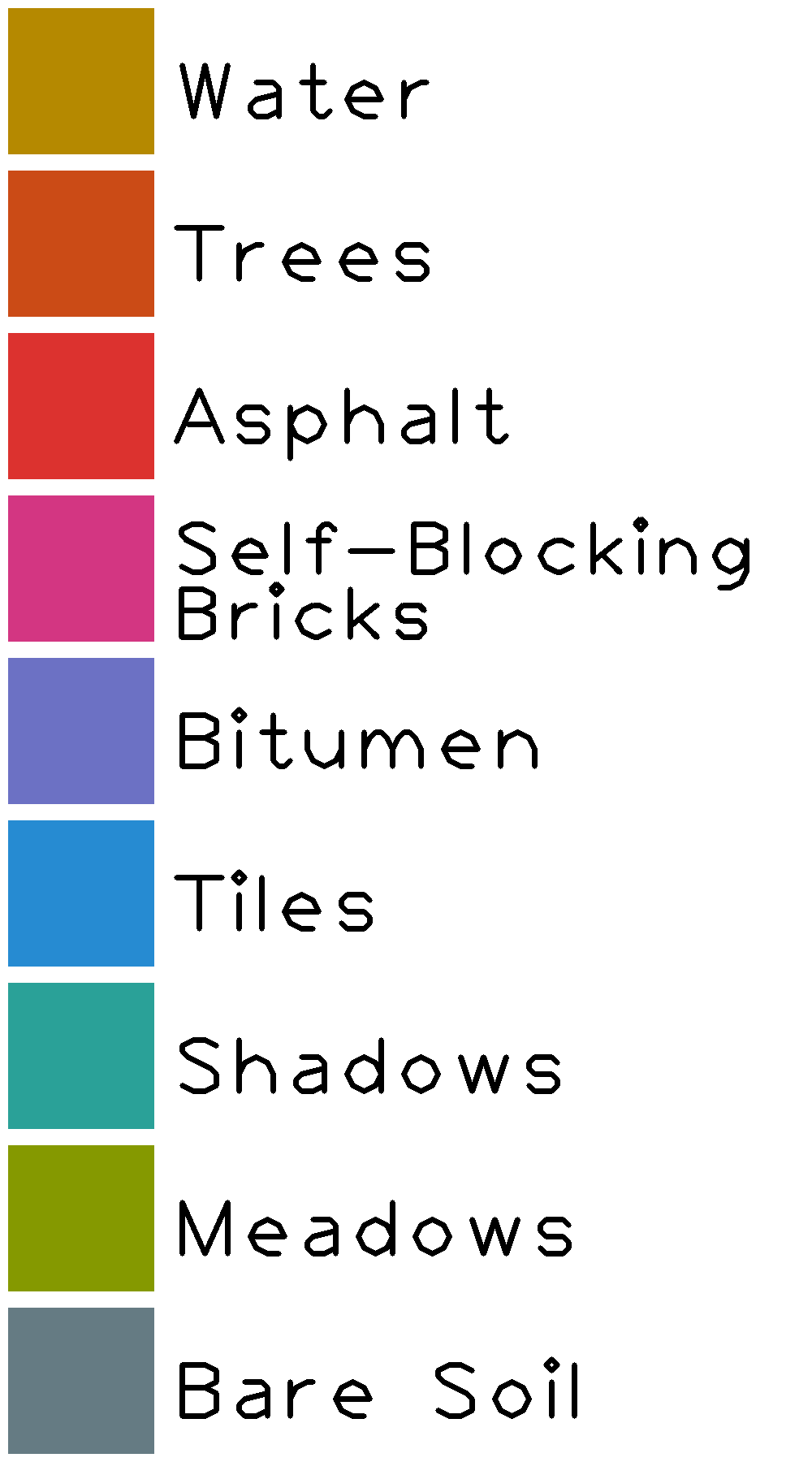}
	\caption{\label{paviac_fc_gt} The false color composite and groundtruth map of Pavia Centre} 
\end{figure}

Three well known hyperspectral images are used in experiments{\footnote{The datasets are available online: \url{http://www.ehu.eus/ccwintco/index.php?title=Hyperspectral_Remote_Sensing_Scenes}}}. The first one is Pavia Centre scene acquired by the Reflective Optics System Imaging Spectrometer (ROSIS) sensor over Pavia, northern Italy. After removing a strip without information, a sub-image of size $1096\times 715$ is taken, with a spatial resolution of $1.3$ meters. After discarding the noisy bands, $L=102$ out of $115$ spectral bands are utilized for analysis, covering a wavelength range from $0.43$ to $0.86$$\mu m$. The groundtruth information consists of $K=9$ classes, as given in \tablename~\ref{label_paviac}.
The false color composite and the groundtruth map of Pavia Center are shown in~\figurename~\ref{paviac_fc_gt}. 

\begin{table}[!hbp] \caption{\label{label_paviac}Reference classes and sizes of training and testing sets of Pavia Centre image}
	\centering
		\begin{tabular}{c|c|c|c|c}
			\hline
			\hline
			No. 	&Class	&Cardinality	&Train	&Test	\\
			\hline
			$1$	&Water	&$65971$	&$200$	&$65771$	\\
			$2$	&Trees	&$7598$	&$200$	&$7398$	\\
			$3$	&Asphalt	&$3090$	&$200$	&$2890$	\\
			$4$	&Self-Blocking	Bricks	&$2685$	&$200$	&$2485$	\\
			$5$	&Bitumen	&$6584$	&$200$	&$6384$	\\
			$6$	&Tiles	&$9248$	&$200$	&$9048$	\\
			$7$	&Shadows	&$7287$	&$200$	&$7087$	\\
			$8$	&Meadows	&$42826$	&$200$	&$42626$	\\
			$9$	&Bare	Soil	&$2863$	&$200$	&$2663$	\\
       \hline
	   \multicolumn{2}{c|}{Total}&148152&1800&146352	\\
        \hline
		\hline
    	\end{tabular}
\end{table}

\begin{figure}
\centering
\graphicspath{{Figures/}}
\emph{\includegraphics[trim = 25mm 30mm 31mm 25mm, clip,width=.12\textwidth]  {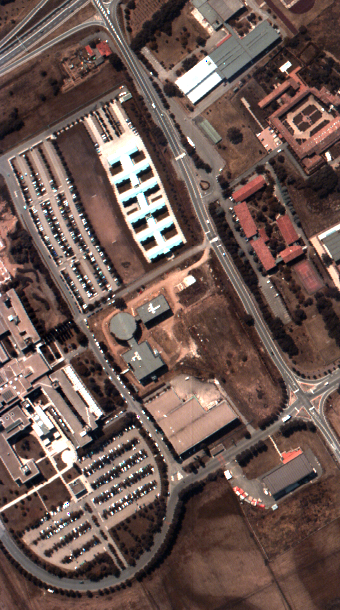}
\includegraphics[trim = 25mm 30mm 31mm 25mm,clip,width=.12\textwidth]  {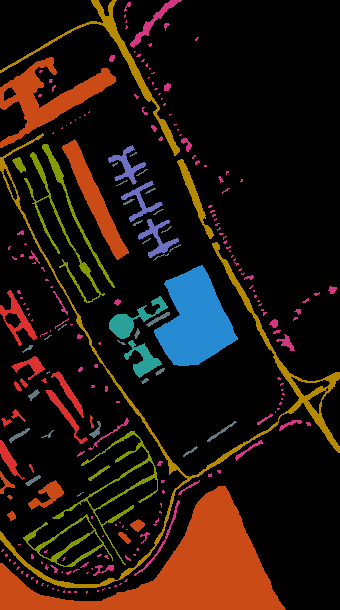}
}\includegraphics[trim = 0mm 0mm 0mm 0mm,clip,width=.11\textwidth]  {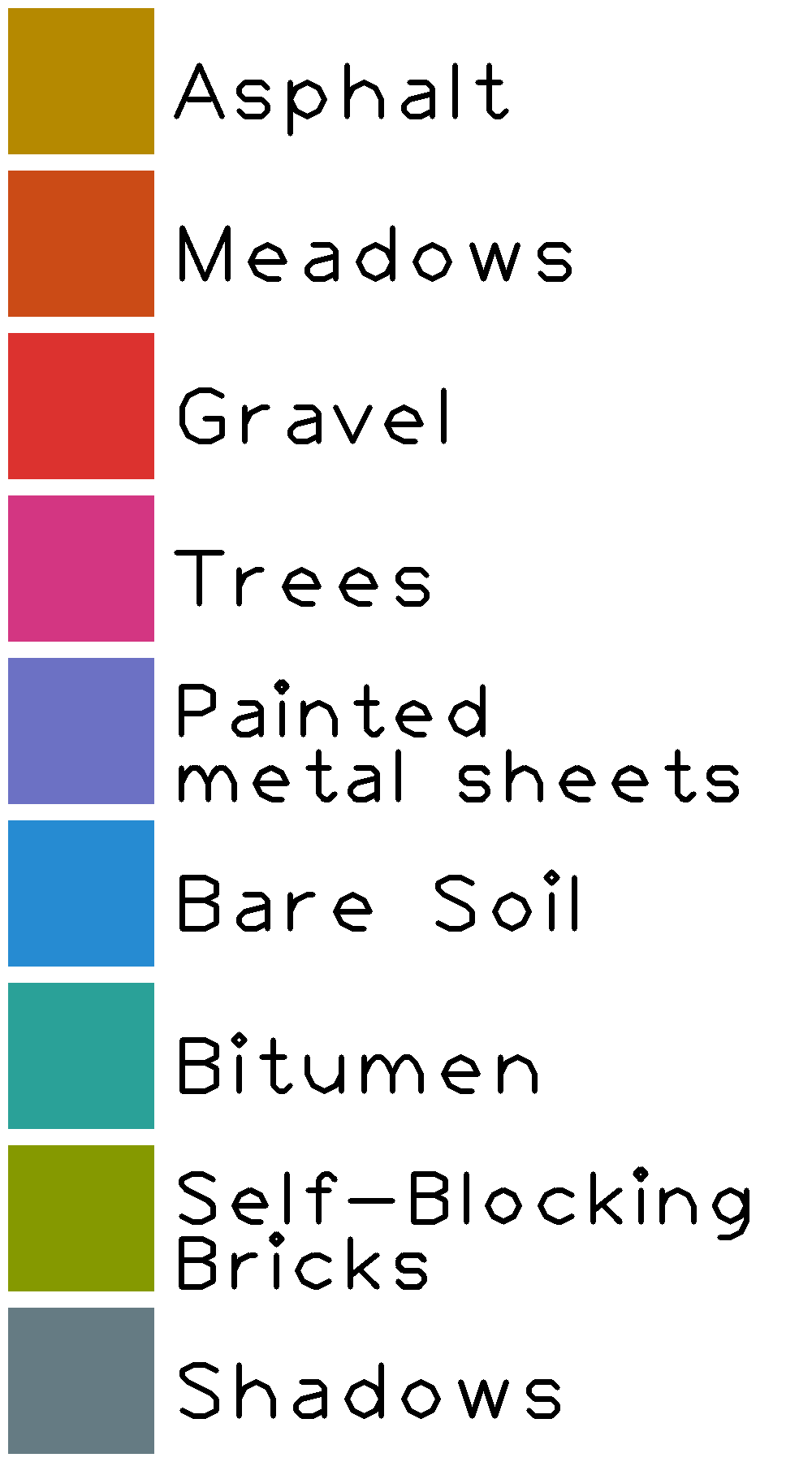}
\caption{\label{paviau_fc_gt} The false color composite and groundtruth representation of Pavia University} 
\end{figure}

The second image is Pavia University scene that is also returned by the ROSIS sensor. This data comprises $610\times 340$ pixels with $L=103$ relatively clean spectral bands of interest. As detailed in~\tablename~ \ref{label_paviau}, the available groundtruth information consists of $K=9$ classes. The false color composite and the groundtruth map are illustrated in~\figurename~\ref{paviau_fc_gt}.
\begin{table}\caption{\label{label_paviau}Reference classes and sizes of training and testing sets of Pavia University image}
	\centering
		\begin{tabular}{c|c|c|c|c}
			\hline
			\hline
			No. 	&Class	&Cardinality	&Train	&Test\\
			\hline
			$1$	&Asphalt	&$6631$	&$200$	&$6431$	\\
			$2$	&Meadows	&$18649$	&$200$	&$18449$	\\
			$3$	&Gravel	&$2099$	&$200$	&$1899$	\\
			$4$	&Trees	&$3064$	&$200$	&$2864$	\\
			$5$	&Painted	metal	sheets	&$1345$	&$200$	&$1145$	\\
			$6$	&Bare	Soil	&$5029$	&$200$	&$4829$	\\
			$7$	&Bitumen	&$1330$	&$200$	&$1130$	\\
			$8$	&Self-Blocking	Bricks	&$3682$	&$200$	&$3482$	\\
			$9$	&Shadows	&$947$	&$200$	&$747$	\\
       \hline
	   \multicolumn{2}{c|}{Total}&42776&1800&40976	\\
			\hline
			\hline
		\end{tabular}
\end{table}

\begin{figure}
	\centering
\graphicspath{{Figures/}}
\includegraphics[trim = 2mm 22mm 13mm 12mm, clip,width=.12\textwidth]  {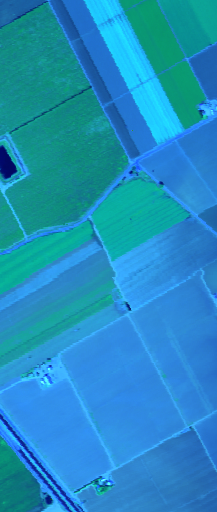}
\includegraphics[trim =  2mm 22mm 13mm 12mm,clip,width=.12\textwidth]  {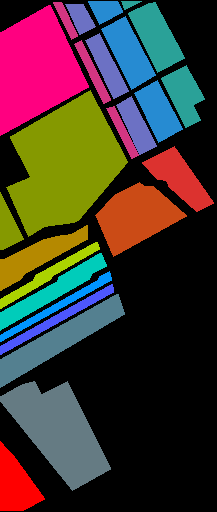}
\includegraphics[trim =  0mm 0mm 0mm 0mm,clip,width=.1\textwidth]  {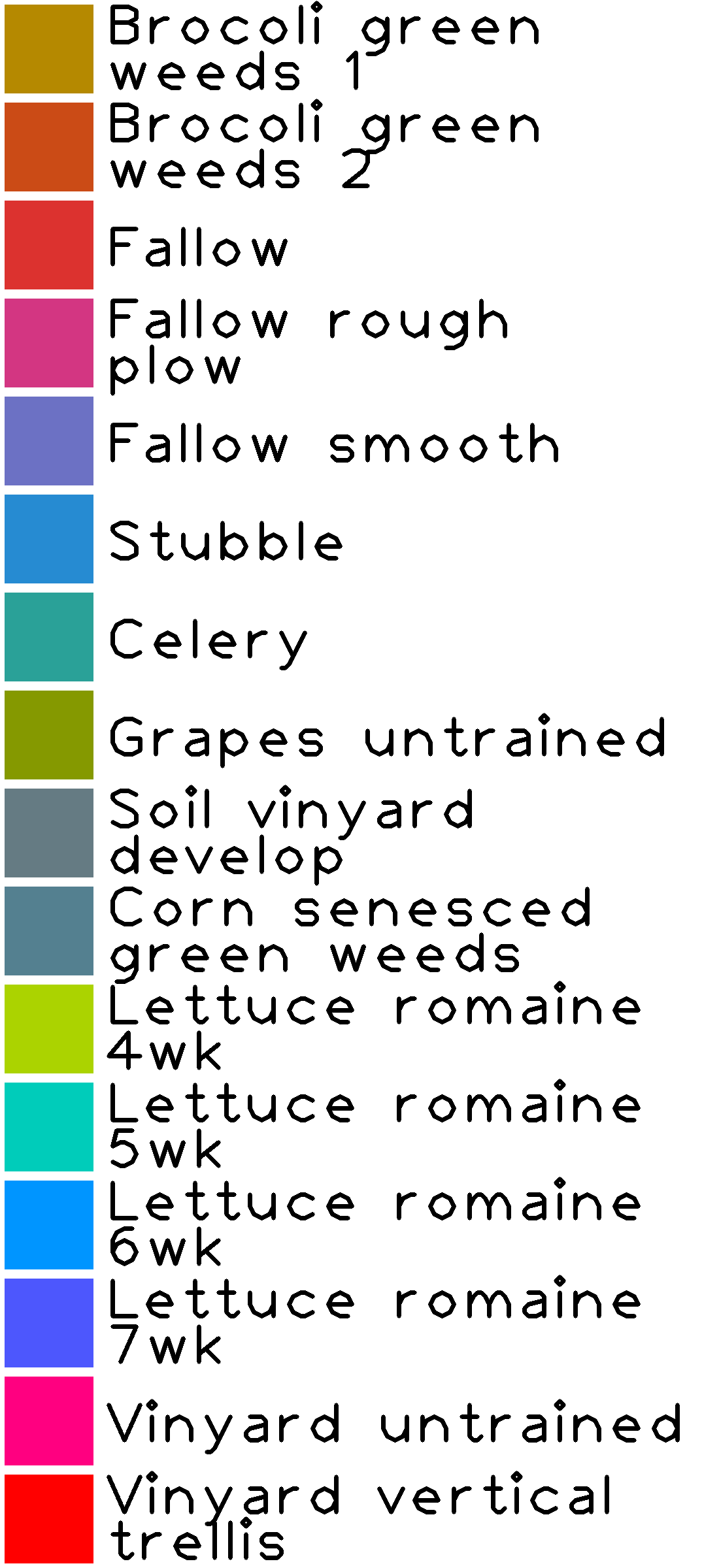}	
	\caption{\label{salinas_fc_gt} The false color composite and groundtruth representation of Salinas} 
\end{figure}

The last image under study is the Salinas scene, collected by the Airborne Visible Infrared Imaging Spectrometer (AVIRIS) over Salinas Valley, California. This image contains $512\times 217$ pixels, characterized by a spatial resolution of $3.7$ meters. After removing the water absorption bands, $L=204$ out of $224$ bands are retained for analysis, with a wavelength range of $0.4-2.5\mu m$. There are $K=16$ classes of land-cover in the groundtruth information, as reported in \tablename~\ref{label_salinas}. The false color composite and the groundtruth map are shown in~\figurename~\ref{salinas_fc_gt}.
\begin{table}\caption{\label{label_salinas}Reference classes and sizes of training and testing sets of Salinas image}
	\centering
		\begin{tabular}{c|c|c|c|c}
			\hline
			\hline
            No. 	&Class	&Cardinality	&Train&   Test	\\
			\hline
			$1$	&Brocoli	green	weeds	1	&$2009$	&$200$	&$1809$	\\
			$2$	&Brocoli	green	weeds	2	&$3726$	&$200$	&$3526$	\\
			$3$	&Fallow	&$1976$	&$200$	&$1776$	\\
			$4$	&Fallow	rough	plow	&$1394$	&$200$	&$1194$	\\
			$5$	&Fallow	smooth	&$2678$	&$200$	&$2478$	\\
			$6$	&Stubble	&$3959$	&$200$	&$3759$	\\
			$7$	&Celery	&$3579$	&$200$	&$3379$	\\
			$8$	&Grapes	untrained	&$11271$	&$200$	&$11071$	\\
			$9$	&Soil	vinyard	develop	&$6203$	&$200$	&$6003$	\\
			$10$	&Corn	senesced	green	weeds	&$3278$	&$200$	&$3078$	\\
			$11$	&Lettuce	romaine	4wk	&$1068$	&$200$	&$868$	\\
			$12$	&Lettuce	romaine	5wk	&$1927$	&$200$	&$1727$	\\
			$13$	&Lettuce	romaine	6wk	&$916$	&$200$	&$716$	\\
			$14$	&Lettuce	romaine	7wk	&$1070$	&$200$	&$870$	\\
			$15$	&Vinyard	untrained	&$7268$	&$200$	&$7068$	\\
			$16$	&Vinyard	vertical	trellis	&$1807$	&$200$	&$1607$	\\
			\hline
	   \multicolumn{2}{c|}{Total}&54129&3200&50929	\\
       \hline
       \hline
		\end{tabular}
\end{table}

Three metrics, {\em i.e.}, overall accuracy (OA), average accuracy (AA), and Kappa coefficient ($\kappa$) are adopted to quantitatively evaluate the classification performance. OA represents the general percentage of the testing samples that are correctly classified, AA computes the average percentage of correctly classified testing samples for each class, and $\kappa$ measures the percentage of classified samples adjusted by the number of agreements that would be expected by chance alone.

\subsection{Experimental setting}
1) \textit{Training and Testing Sets:}
Before subsequent training and testing procedures, each raw image is firstly normalized to have zero mean and unit variance. 
To alleviate overfitting caused by limited training samples, we enlarge the training set by virtual samples generation in a similar way as did in~\cite{chen2016deep}. Let $\bx_1$ and $\bx_2$ be two samples from the same class, a virtual sample $y$ with the same label is generated by $\by = q \bx_1+(1-q) \bx_2$,
where the range of $q$ is empirically set as $[-1,2]$.
In the following, the training set of each class consists of 200 real samples enhanced by 80000 virtual ones. All the remaining samples with groundtruth information are used for testing. 

2) \textit{Network Configuration:}
The same network structure, as illustrated in~\figurename~\ref{struct}, is utilized on each of the datasets,
except for various sizes of input and output layers that are determined case by case. The mini-batch size is 512, and the learning rate in SGD is initialized as $0.01$, and decays every $20000$ mini-batches with a multiplier $\sqrt{0.1}\approx 0.3162$.
Dropout strategy is adopted in Layer $3$ with the dropout ratio to be $0.3$. The proposed framework is built on the open source deep learning framework Caffe~\cite{jia2014caffe}, and each experiment is repeated $5$ times.
\subsection{Investigation of spectrum-based FE and classification}

1) \textit{Enhancing Discriminative Ability with Center Loss:}
We analyze how does center loss enhance the discriminative ability on spectral feature extraction. To this end, experiments are performed on Pavia Center, using two neuron networks supervised by conventional softmax loss (ANN) and by a joint loss combining softmax loss and center loss (ANNC), respectively. For fair comparison, all the other conditions are identically set in both cases, such as network structure, training set, hyperparameters, {\em etc}.

\begin{figure}
	\centering
	\graphicspath{{Figures/}}
\subfigure[ANNC]{
		\includegraphics[trim = 4mm 1mm 2mm 4mm, clip,width=0.4\textwidth]  {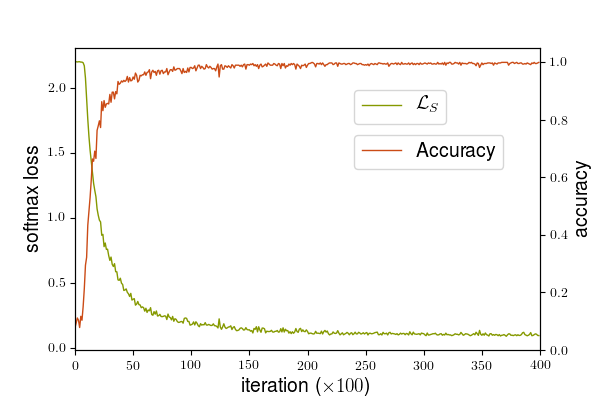}}
\subfigure[ANN]{
	    \includegraphics[trim = 4mm 1mm 2mm 4mm, clip,width=0.4\textwidth]  {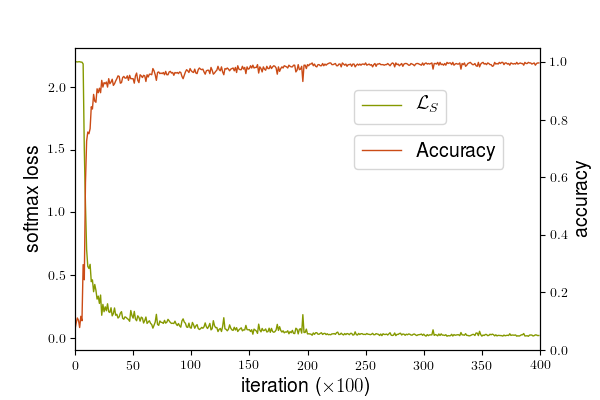}}
	\caption{\label{loss_accu}Comparison of (a) ANNC and (b) ANN in terms of softmax loss $\mathcal{L}_S$ and classification accuracy along with iterations ($\times 100$).}
\end{figure}

The behaviours of softmax loss, {\em i.e.} $\mathcal{L}_S$ in \eqref{eq:Softmax}, and classification accuracy of models ANN and ANNC in training stage are illustrated in~\figurename~\ref{loss_accu}. We observe that training with center loss leads to
more stable changes in terms of both softmax loss and classification accuracy along with iterations, when compared to its counterpart without center loss.

\begin{figure}
\centering
\graphicspath{{Figures/}}
\subfigure[ANNC]{
		\includegraphics[trim = 4mm 1mm 2mm 4mm, clip,width=0.4\textwidth]  {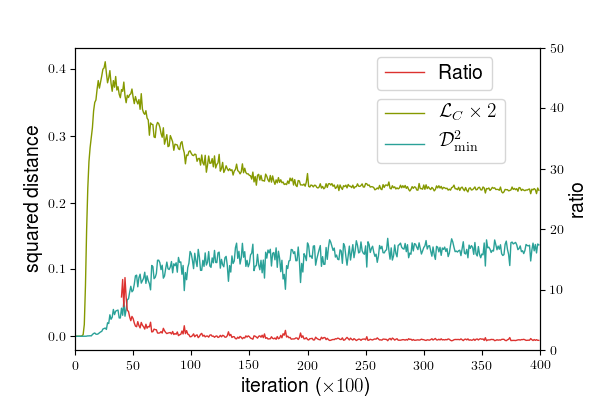}}
\subfigure[ANNC]{
		\includegraphics[trim = 4mm 1mm 2mm 4mm, clip,width=0.4\textwidth]  {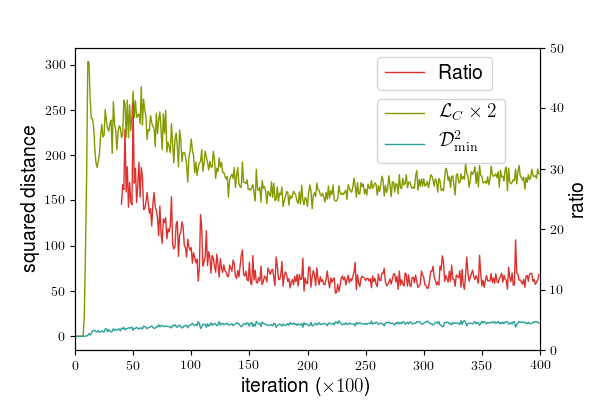}
	}
	\caption{\label{center_dist}Comparison of (a) ANNC and (b) ANN in terms of center loss $2\times\mathcal{L}_C$, $\mathcal{D}_{min}^2$ and the ratio between them, along with iterations ($\times 100$).}
\end{figure}

We investigate how the distribution of features changes in Euclidean space at the training stage. To this end, three statistics are monitored, as shown in \figurename~\ref{center_dist}. The center loss, {\em i.e.} $\mathcal{L}_C$ in \eqref{center_loss}, measures the averaged intra-class distance of features among classes, with a smaller value reflecting better intra-class compactness. As a measurement of inter-class variation, the minimum squared distance between any two centers are traced, denoted by $\mathcal{D}^2_{\mathrm{min}}$. The ratio between these two statistics are also considered, with a smaller magnitude signifying better discriminative ability.

We observe that training with center loss yields smaller $\mathcal{L}_C$ and greater $\mathcal{D}^2_{\mathrm{min}}$, with the ratio between them several times smaller, when compared to its ANN counterpart. This demonstrates that the learned features under ANNC are more gathered within class while different classes are more distant. Moreover, the curves of ANNC are smoother, showing more steady convergence property over iterations.

\begin{figure}
	\centering
	\graphicspath{{Figures/}}
\subfigure[Features under ANNC]{
	    \includegraphics[trim = 12mm 5mm 11mm 0mm, clip,width=0.48\textwidth]  {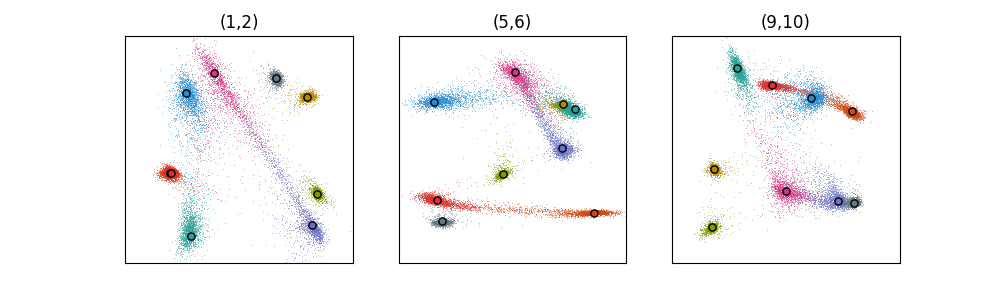}}
\subfigure[Features under ANN]{
	    \includegraphics[trim = 12mm 5mm 11mm 0mm, clip,width=0.48\textwidth]  {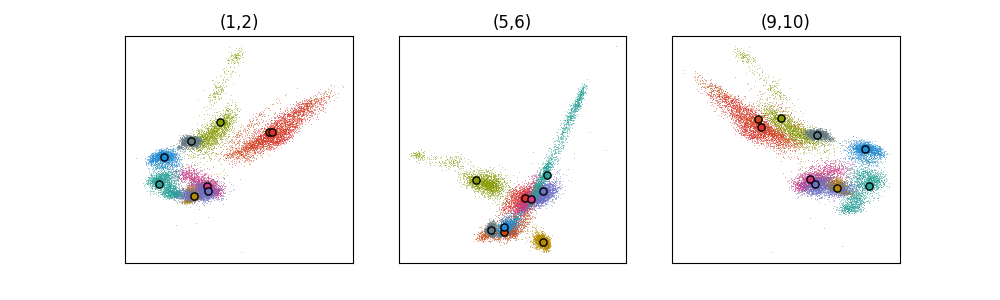}}
	\caption{\label{features}The distributions of spectral features on training samples, using networks ANNC and ANN. Features from various classes are denoted in different colors. }
\end{figure}

\figurename~\ref{features} visualises the spatial distribution of spectral features extracted by ANN and ANNC, on $2000$ random testing samples. For illustration purpose, the $32$-dimension features are projected to certain $2$ coordinates $(i,j)$. As observed, the features extracted by ANNC are gathered more closely within class while the distances between different classes are more enlarged, compared to the case ANN.
The above discussions validate that training with center loss helps to improve the discriminative property of the network that is otherwise trained by trivial softmax loss.

2) \textit{Classification with Spectral Features:}
To demonstrate the effectiveness of the discriminative spectral features for subsequent classification, comparative experiments are performed on spectral features that are extracted with ANN and ANNC, respectively.
Two classifiers, namely the aforementioned center classifier (CC) and the logistic regression (LR) are considered. The results on datasets Pavia Center, Pavia University and Salinas are presented in Tables~\ref{result_paviac}-\ref{result_salinas}.

Compared with ANN, the spectral features extracted by ANNC generally leads to better classification accuracy in terms of AA, OA and $\kappa$, using either classifier on all the datasets. It demonstrates that the discriminative ability induced by center loss has a positive influence on classification performance.
To be precise, ANNC combined with CC yields best classification accuracy on Pavia University and Salinas. Regarding Pavia Center, it is ANNC combined with LR that archives the best performance, second by ANNC classified with CC.
The combination of ANN and CC leads to the worst results on all the datasets, since features trained with ANN are lack of intra class compactness and inter class dispersion, thus not suitable for CC.

\begin{figure}
	\graphicspath{{Figures/}}
	\centering
	    \graphicspath{{Figures/}}
	   \includegraphics[trim = 2mm 22mm 13mm 12mm, clip,width=.12\textwidth]{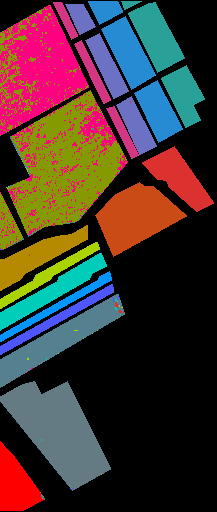}
	   \includegraphics[trim =  2mm 22mm 13mm 12mm,clip,width=.12\textwidth]{salinas_groundtruth.png}
	\caption{\label{salias_pred_gt} Comparison of predicted labels by ANN and groundtruth on Salinas. As observed, most failed predictions are caused by the mislabeling
between classes Grapes untrained and Vinyard untrained located at up-left corner.}
\end{figure}

It is noteworthy that ANNC is effective in addressing the confusion between similar classes.
Concerning Salinas, while most classes are well recognized with high classification accuracy achieved by both ANN and ANNC, there are two classes, namely Grapes untrained and Vinyard untrained, that are locally adjacent and hard to be distinguished one from the other. By examining the predictions and groundtruth labels, one could find that most of failed predictions are caused by the mislabeling between these two classes located at the up-left region of the image, see \figurename~\ref{salias_pred_gt}. \tablename ~\ref{result_salinas} illustrates that ANNC improves the classification result on such difficult task, compared with its ANN counterpart, using either classifier.

\begin{table}\caption{\label{result_paviac}The results on Pavia Centre scene}
    \centering
    	\begin{tabular}{c|c|c|c|c}
    		\hline \hline
			Classifier& \multicolumn{2}{c|}{LR} & \multicolumn{2}{c}{CC} \\ \hline
            FE method & ANN        & ANNC       & ANN        & ANNC       \\ \hline \hline
    		Water                & 99.92      & 99.96      & 99.94      & 99.96      \\ \cline{2-5}
    		Trees                & 96.13      & 97.10      & 90.61      & 97.24      \\ \cline{2-5}
    		Asphalt              & 97.38      & 97.21      & 92.33      & 97.11      \\ \cline{2-5}
    		Self-Blocking Bricks & 97.15      & 96.61      & 94.13      & 96.60      \\ \cline{2-5}
    		Bitumen              & 95.65      & 95.67      & 93.50      & 95.60      \\ \cline{2-5}
    		Tiles                & 97.94      & 98.10      & 99.32      & 98.22      \\ \cline{2-5}
    		Shadows              & 94.05      & 94.34      & 88.84      & 94.29      \\ \cline{2-5}
    		Meadows              & 99.37      & 99.44      & 94.93      & 99.37      \\ \cline{2-5}
    		Bare Soil            & 99.92      & 99.95      & 99.99      & 99.94      \\ \hline \hline
    		OA$(\%)$                   & 98.88      & \!\!\!\!\!\textcircled{\tiny1}{98.98}      & 96.90   & \!\!\!\!\!\textcircled{\tiny2}98.96 \\ \cline{2-5}
    		AA$(\%)$                   & 97.50      & \!\!\!\!\!\textcircled{\tiny1}{97.60}      & 94.84   & \!\!\!\!\!\textcircled{\tiny2}97.59 \\ \cline{2-5}
    		$\kappa$     & 0.9840     &  \!\!\textcircled{\tiny1}{0.9855}   & 0.9561     & \!\!\textcircled{\tiny2}0.9852 \\ \hline \hline
    	\end{tabular}
\end{table}

\begin{table}\caption{\label{result_paviau}The results on Pavia University scene}
	\centering
		\begin{tabular}{c|c|c|c|c}
			\hline \hline
			Classifier& \multicolumn{2}{c|}{LR} & \multicolumn{2}{c}{CC} \\ \hline
			FE method & ANN        & ANNC       & ANN        & ANNC       \\ \hline \hline
			Asphalt              & 88.41      & 90.06      & 88.46      & 90.47      \\ \cline{2-5}
			Meadows              & 94.12      & 95.27      & 58.21      & 94.94      \\ \cline{2-5}
			Gravel               & 85.76      & 87.71      & 84.09      & 85.52      \\ \cline{2-5}
			Trees                & 97.43      & 97.35      & 90.77      & 97.41      \\ \cline{2-5}
			Painted metal sheets & 99.96      & 99.98      & 99.05      & 99.98      \\ \cline{2-5}
			Bare Soil            & 95.68      & 95.71      & 90.01      & 96.09      \\ \cline{2-5}
			Bitumen              & 95.41      & 95.07      & 95.61      & 94.96      \\ \cline{2-5}
			Self-Blocking Bricks & 88.57      & 85.30      & 88.85      & 88.06      \\ \cline{2-5}
			Shadows              & 99.70      & 99.81      & 99.86      & 99.81      \\ \hline \hline
			OA$(\%)$                   & 93.08      & \!\!\!\!\!\textcircled{\tiny2}93.66      & 75.69      & \!\!\!\!\!\textcircled{\tiny1}{93.76}    \\ \cline{2-5}
			AA$(\%)$                   & 93.89      & \!\!\!\!\!\textcircled{\tiny2}94.03      & 88.32      & \!\!\!\!\!\textcircled{\tiny1}{94.14}      \\ \cline{2-5}
			$\kappa$                & 0.9078     & \!\!\textcircled{\tiny2}0.9153 & 0.6976     & \!\!\textcircled{\tiny1}{0.9166}     \\ \hline \hline
		\end{tabular}
\end{table}

\begin{table}\caption{\label{result_salinas}The results on Salinas scene}
	\centering
		\begin{tabular}{c|c|c|c|c}
			\hline \hline
			\multicolumn{1}{c|}{Classifier} & \multicolumn{2}{c|}{LR} & \multicolumn{2}{c}{CC} \\ \hline
			\multicolumn{1}{c|}{FE method}  & ANN        & ANNC       & ANN        & ANNC       \\ \hline \hline
			Brocoli\_green\_weeds\_1         & 99.91      & 99.81      & 99.96      & 99.80      \\ \cline{2-5}
			Brocoli\_green\_weeds\_2         & 99.91      & 99.97      & 99.54      & 99.96      \\ \cline{2-5}
			Fallow                           & 99.68      & 99.76      & 95.23      & 99.86      \\ \cline{2-5}
			Fallow\_rough\_plow              & 99.66      & 99.66      & 99.71      & 99.68      \\ \cline{2-5}
			Fallow\_smooth                   & 98.97      & 99.10      & 98.70      & 99.08      \\ \cline{2-5}
			Stubble                          & 99.92      & 99.98      & 98.58      & 99.99      \\ \cline{2-5}
			Celery                           & 99.84      & 99.85      & 99.90      & 99.85      \\ \cline{2-5}
			\textbf{Grapes\_untrained}                & 79.33      & 80.52      & 79.27      & 83.26      \\ \cline{2-5}
			Soil\_vinyard\_develop           & 99.77      & 99.69      & 99.89      & 99.70      \\ \cline{2-5}
			Corn\_senesced\_green\_weeds     & 96.03      & 96.47      & 88.94      & 96.47      \\ \cline{2-5}
			Lettuce\_romaine\_4wk            & 99.51      & 99.74      & 97.74      & 99.77      \\ \cline{2-5}
			Lettuce\_romaine\_5wk            & 100.00     & 100.00     & 100.00     & 100.00     \\ \cline{2-5}
			Lettuce\_romaine\_6wk            & 99.77      & 99.58      & 99.60      & 99.58      \\ \cline{2-5}
			Lettuce\_romaine\_7wk            & 99.21      & 99.35      & 97.70      & 99.35      \\ \cline{2-5}
			\textbf{Vinyard\_untrained }              & 81.96      & 82.75      & 68.17      & 80.14      \\ \cline{2-5}
			Vinyard\_vertical\_trellis       & 99.29      & 99.06      & 97.82      & 99.09      \\ \hline \hline
			OA$(\%)$   & 92.58      & \!\!\!\!\!\textcircled{\tiny2}92.98      & 89.84      & \!\!\!\!\!\textcircled{\tiny1}93.22   \\ \cline{2-5}
			AA$(\%)$   & 97.05      & \!\!\!\!\!\textcircled{\tiny2}97.21      & 95.05      & \!\!\!\!\!\textcircled{\tiny1}97.22       \\ \cline{2-5}
			$\kappa$   & 0.9170     & \!\!\textcircled{\tiny2}0.9215     & 0.8863     & \!\!\textcircled{\tiny1}0.9241  \\ \hline \hline
		\end{tabular}
\end{table}

\subsection{Performance of Adaptive Spectral-Spatial Center Classifier (ANNC-ASSCC)}

The performance of ANNC-ASSCC is compared with two state-of-the-art methods, namely PPF~\cite{li2017hyperspectral} and MASR~\cite{Fang2014Spectral}. To examine the proposed spatial integration strategy, ANNC-SCC is also considered. As shown in Tables~\ref{av_paviac}-\ref{av_salinas}, ANNC-ASSCC achieves classification improvements over ANNC-SCC in terms of all the metrics on three datasets. It demonstrates that the  discriminative features facilitate the postprocessing, and the proposed spatial integration strategy is effective in classifying these features.

We also observe that ANNC-ASSCC outperforms the comparative methods on Pavia Center and Pavia Unversity, but is inferior to MASR on Salinas. One possible reason is that MASR exploits multiscale neighborhoods at the testing stage without removing the training samples.
To guarantee a fair comparison and clarify how does train-test overlap influence the classification performance, supplementary experiments are performed with slight modifications on MASR.
At the testing stage, we design MASR-t that exclusively deteriorates training information by replacing all the training pixels with a $L$-dimensional vector of identical entries 0.01. As a contrast, MASR-r introduces an even disturb by randomly replacing the same number of pixels with the previous vector. The results are appended to~\tablename~\ref{av_salinas}. Compared with MASR, MASR-r barely deteriorates the classification performance with slight drops less than 0.1\% in terms of OA, AA and $\kappa$. However, MASR-t yields over 2\% decreases in terms of OA and $\kappa$, and only achieves comparable results as the proposed ANNC-ASSCC.
To conclude, the proposed method provides competitive classification results over the state-of-the-art approaches on three datasets.

\begin{table*}[!htb]
	\centering
		\caption{\label{av_paviac}Classification accuracy (averaged over 5 runs) obtained by MASR~\cite{Fang2014Spectral}, PPF~\cite{li2017hyperspectral}, ANNC-SCC and ANNC-ASSCC on Pavia Centre}
		\begin{tabular}{c|c|c|c|c}
		\hline
        \hline
            {} 					   & MASR   & PPF    & ANNC-SCC & ANNC-ASSCC \\ \hline
        Water                & 99.87  & 99.15  & 99.91   & 100.00      \\ \cline{2-5}
        Trees                & 94.22  & 97.96  & 96.34   & 98.75       \\ \cline{2-5}
        Asphalt              & 99.45  & 97.37  & 97.72   & 99.26       \\ \cline{2-5}
        Self-Blocking Bricks & 99.98  & 99.27  & 96.67   & 99.96       \\ \cline{2-5}
        Bitumen              & 98.75  & 98.79  & 95.94   & 99.35       \\ \cline{2-5}
        Tiles                & 80.23  & 98.95  & 98.28   & 99.73       \\ \cline{2-5}
        Shadows              & 99.34  & 94.36  & 93.62   & 97.49       \\ \cline{2-5}
        Meadows              & 99.90  & 99.90  & 99.41   & 99.99       \\ \cline{2-5}
        Bare Soil            & 84.80  & 99.96  & 99.95   & 98.72       \\ \hline \hline
        OA($\%$)                   & 98.02  &\!\!\!\!\!\textcircled{\tiny2}99.03  & 98.91   &  \!\!\!\!\!\textcircled{\tiny1}99.73       \\ \cline{2-5}
        AA($\%$)                   & 95.17  &\!\!\!\!\!\textcircled{\tiny2}98.41  & 97.54   &  \!\!\!\!\!\textcircled{\tiny1}99.25       \\ \cline{2-5}
        $\kappa$                   &0.9719  &\!\!\textcircled{\tiny2}0.9862    & 0.9845  & \!\!\textcircled{\tiny1}0.9961      \\ \hline \hline
		\end{tabular}
\end{table*}
\begin{table*}
	\centering
	\caption{\label{av_paviau}Classification accuracy (averaged over 5 runs) obtained by MASR~\cite{Fang2014Spectral}, PPF~\cite{li2017hyperspectral}, ANNC-SCC and ANNC-ASSCC on Pavia University}
	\begin{tabular}{c|c|c|c|c}
		\hline
		\hline
		{} 					   & MASR   & PPF    & ANNC-SCC & ANNC-ASSCC \\ \hline
        Asphalt              & 89.97  & 97.25  & 90.06   & 98.69       \\ \cline{2-5}
        Meadows              & 98.78  & 95.24  & 95.75   & 99.97       \\ \cline{2-5}
        Gravel               & 99.78  & 94.17  & 86.76   & 93.85       \\ \cline{2-5}
        Trees                & 97.47  & 97.20  & 97.54   & 96.68       \\ \cline{2-5}
        Painted metal sheets & 100.00 & 100.00 & 100.00  & 100.00      \\ \cline{2-5}
        Bare Soil            & 99.87  & 99.37  & 94.88   & 100.00      \\ \cline{2-5}
        Bitumen              & 100.00 & 96.16  & 94.81   & 96.88       \\ \cline{2-5}
        Self-Blocking Bricks & 98.76  & 93.83  & 83.68   & 93.11       \\ \cline{2-5}
        Shadows              & 92.17  & 99.46  & 99.63   & 97.62       \\ \hline \hline
        OA($\%$)                   & \!\!\!\!\!\textcircled{\tiny2}97.42  & 96.25  & 93.60   & \!\!\!\!\!\textcircled{\tiny1}98.55       \\ \cline{2-5}
        AA($\%$)                   & \!\!\!\!\!\textcircled{\tiny2}97.42  & 96.97  & 93.68   & \!\!\!\!\!\textcircled{\tiny1}97.42       \\ \cline{2-5}
        $\kappa$                & \!\!\textcircled{\tiny2}0.9654 & 0.9499 & 0.9145  &    \!\!\textcircled{\tiny1}0.9805   \\ \hline \hline
	\end{tabular}	
\end{table*}
\begin{table*}
	\centering
		\caption{\label{av_salinas}Classification accuracy (averaged over 5 runs) obtained by MASR~\cite{Fang2014Spectral}, PPF~\cite{li2017hyperspectral}, ANNC-SCC and ANNC-ASSCC on Salinas}
		\begin{tabular}{c|c|c|c|c||c|c}
		\hline
        \hline
            {} 					   & MASR   & PPF    & ANNC-SCC & ANNC-ASSCC & MASR-r & MASR-t\\ \hline
			Brocoli\_green\_weeds\_1     & 100.00 & 100.00 & 99.82   & 100.00  & 100.00   & 100.00        \\ \cline{2-7}
			Brocoli\_green\_weeds\_2     & 99.97  & 99.89  & 99.91   & 100.00  & 99.89    & 99.86        \\ \cline{2-7}
			Fallow                       & 100.00 & 99.61  & 99.91   & 100.00  & 100.00   & 100.00        \\ \cline{2-7}
			Fallow\_rough\_plow          & 99.89  & 99.50  & 99.63   & 99.53  & 99.92    & 99.50         \\ \cline{2-7}
			Fallow\_smooth               & 99.58  & 98.35  & 99.10   & 99.67 & 99.80    & 99.44         \\ \cline{2-7}
			Stubble                      & 100.00 & 99.97  & 99.96   & 100.00  & 99.95    & 100.00       \\ \cline{2-7}
			Celery                       & 99.95  & 100.00 & 99.85   & 100.00  & 99.97    & 99.94    \\ \cline{2-7}
			Grapes\_untrained            & 97.82  & 88.68  & 83.95   & 92.34    & 97.12    & 87.20    \\ \cline{2-7}
			Soil\_vinyard\_develop       & 99.99  & 98.33  & 99.78   & 99.99  & 100.00   & 100.00       \\ \cline{2-7}
			Corn\_senesced\_green\_weeds & 99.90  & 98.60  & 95.91   & 99.44   & 99.81    & 99.09     \\ \cline{2-7}
			Lettuce\_romaine\_4wk        & 100.00 & 99.53  & 99.65   & 100.00   & 100.00   & 100.00   \\ \cline{2-7}
			Lettuce\_romaine\_5wk        & 99.98  & 100.00 & 100.00  & 100.00   & 100.00   & 100.00    \\ \cline{2-7}
			Lettuce\_romaine\_6wk        & 100.00 & 99.44  & 99.47   & 99.50   & 100.00   & 100.00    \\ \cline{2-7}
			Lettuce\_romaine\_7wk        & 99.90  & 98.97  & 99.15   & 99.63    & 99.77    & 99.54    \\ \cline{2-7}
			Vinyard\_untrained           & 99.22  & 83.53  & 78.49   & 90.79   & 99.75    & 96.87    \\ \cline{2-7}
			Vinyard\_vertical\_trellis   & 99.99  & 99.32  & 99.30   & 99.99   & 99.94    & 99.88       \\ \hline \hline
			OA($\%$)                           & \!\!\!\!\!\textcircled{\tiny1}99.38  & 94.80  & 93.12   & \!\!\!\!\!\textcircled{\tiny2}96.98  & 99.30    & 96.66      \\
\cline{2-7}
			AA($\%$)                           & \!\!\!\!\!\textcircled{\tiny1}99.76  & 97.73  & 97.12   & \!\!\!\!\!\textcircled{\tiny2}98.81   & 99.74    & 98.83        \\
\cline{2-7}
			$\kappa$                        & \!\!\textcircled{\tiny1}0.9930 & 0.9418 & 0.9230  & \!\!\textcircled{\tiny2}0.9662 & 0.9921   & 0.9627      \\ \hline \hline
		\end{tabular}
\end{table*}

\section{Conclusion}\label{sec: Conclusion}

This paper presented an ANN-based framework for spectral-spatial hyperspectral feature extraction and classification. The center loss was introduced to train the network in order to enhance the discriminative ability of the model. Based on the learned network, two classification strategies were proposed at the testing stage. In ANNC-SCC, the spectral features were classified using the center classifier. In ANNC-ASSCC, multi-scale spatial information was adaptively integrated to the spectral features, and the classification was performed using a voting strategy. The effectiveness of the proposed methods were demonstrated on three well-known hyperspectral images.
As for future work, other loss functions proposed in deep metric learning, such as triplet
loss and magnet loss, deserve investigation to address the hyperspectral feature extraction and classification problems.

\section*{Acknowledgment}

This work was supported by the National Science Foundation for Young Scientists of China under grant 61701337.

\bibliography{AlanGuo,bib_fei}

\end{document}